\documentclass[10pt,twocolumn,letterpaper]{article}
\usepackage{float}
\usepackage{stfloats}
\usepackage{multirow}
\usepackage{graphicx}
\usepackage{colortbl}
\usepackage{array}
\usepackage{tabularray}
\usepackage{comment}
\usepackage{makecell}
\usepackage[pagenumbers]{cvpr} % To force page numbers, e.g. for an arXiv version

\definecolor{cvprblue}{rgb}{0.21,0.49,0.74}
\usepackage[pagebackref,breaklinks,colorlinks,allcolors=cvprblue]{hyperref}

\def\paperID{6487} % *** Enter the Paper ID here
\def\confName{CVPR}
\def\confYear{2025}

\newcommand*{\affaddr}[1]{#1} % No op here. Customize it for different styles.
\newcommand*{\affmark}[1][*]{\textsuperscript{#1}}
\newcommand*{\email}[1]{\small{\texttt{#1}}}

\title{MambaVLT: Time-Evolving Multimodal State Space Model for \\ Vision-Language Tracking}
\author{Xinqi Liu\affmark[1$\dagger$], Li Zhou\affmark[1$\dagger$], Zikun Zhou\affmark[2*], Jianqiu Chen\affmark[1], and Zhenyu He\affmark[1,*]\\\affaddr{\affmark[1]Harbin Institute of Technology, Shenzhen}\quad\affaddr{\affmark[2]Peng Cheng Laboratory}\\
\email{\{xqliu01,lizhou.hit,zhouzikunhit,jianqiuer\}@gmail.com\quad zhenyuhe@hit.edu.cn}\\}
\begin{document}

\maketitle

\renewcommand{\thefootnote}{\fnsymbol{footnote}} 
\footnotetext{$^{\dagger}$Xinqi Liu and Li Zhou contribute equally.~ $^*$Zikun Zhou and Zhenyu He are Corresponding authors.}

\begin{abstract}
% 任务名称如何定义
The vision-language tracking task aims to perform object tracking based on various modality references. Existing Transformer-based vision-language tracking methods have made remarkable progress by leveraging the global modeling ability of self-attention. However, current approaches still face challenges in effectively exploiting the temporal information and dynamically updating reference features during tracking. Recently, the State Space Model (SSM), known as Mamba, has shown astonishing ability in efficient long-sequence modeling. Particularly, its state space evolving process demonstrates promising capabilities in memorizing multimodal temporal information with linear complexity. Witnessing its success, we propose a Mamba-based vision-language tracking model to exploit its state space evolving ability in temporal space for robust multimodal tracking, dubbed MambaVLT. In particular, our approach mainly integrates a time-evolving hybrid state space block and a selective locality enhancement block, to capture contextual information for multimodal modeling and adaptive reference feature update. Besides, we introduce a modality-selection module that dynamically adjusts the weighting between visual and language references, mitigating potential ambiguities from either reference type. Extensive experimental results show that our method performs favorably against state-of-the-art trackers across diverse benchmarks.

% Specially, our approach presents a Long-term Temporal State Space (LTSS) module, integrating a Hybrid Multi-modal State Space (HMSS) block with temporal awareness and a Selective Locality Enhancement (SLE) block. The HMSS block comprises state space evolving mechanism and modality-guided bidirectional scanning to capture context information and dynamically model visual and language information.

% Specially, we present a multimodal state space evolving mechanism with modality-guided bidirectional scan to effectively model and update features of visual and language information. 

\end{abstract}    
\section{Introduction}
\label{sec:intro}

% 在基于bounding box的追踪中，目标根据

\begin{figure}[t]
    \centering
    \includegraphics[width=1\linewidth]{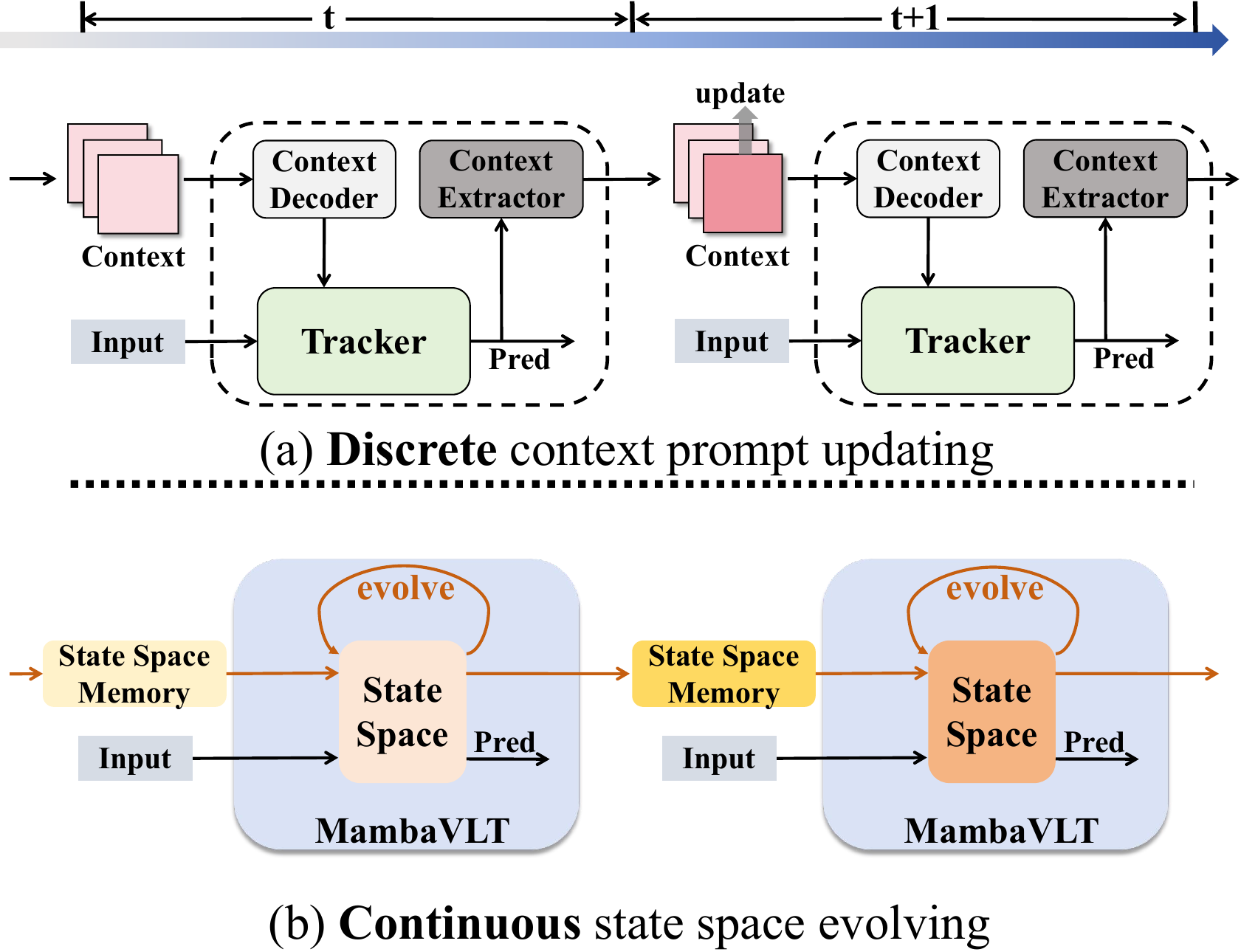}
    \caption{Illustration of two ways for capturing temporal context information. (a) Vision-language tracker with discrete context prompt. (b) Our MambaVLT with continuous time-evolving state space for temporal information transmission.}
    \label{fig:pipline}
\end{figure}

% context-aware并不是对文本信息进行更新，更新可能包含删除，但也包含添加，context-ware中只有对文本信息的删除，算是文本的调制。本身文本之后的pad也会参与信息传递，会被用作文本的更新
Single object tracking involves localizing a target in a video based on provided reference information, which may be an initial bounding box~\cite{yilmaz2006object}, a language specification~\cite{li2017tracking}, or a combination of both~\cite{li2017tracking}. This technology has diverse applications, including video surveillance, robotics, and autonomous vehicles. Tracking by initial bounding box~\cite{ye2022joint,lin2022swintrack,gao2023generalized} is an extensively studied tracking task. A common solution is cropping a template based on the bounding box in the first frame as a reference and accordingly locating the target in subsequent frames. Nonetheless, solely relying on the visual template without direct semantics may lead to ambiguity~\cite{wang2021towards}. Recently, tracking by natural language specification~\cite{li2017tracking} and tracking by both language and box specification~\cite{li2017tracking} have been proposed to address this issue. These approaches incorporate language descriptions as references, facilitating natural human-computer interaction. Previous studies~\cite{zhou2023joint,shao2024context,ma2024unifying} have made significant progress under various reference settings. However, they are still constrained by their limited ability to capture long-term temporal information and adaptively update the reference information as the tracker operates on a video. 

Typically, the appearance and motion mode of the target keep varying in the video. Previous works~\cite{zhou2023joint,shao2024context,ma2024unifying} have introduced different methods to adapt to these temporal variations, which can be summarized as a discrete approach to extracting contextual features, as shown in Figure \ref{fig:pipline}(a). It can be divided into two steps: (1) generating a context prompt by a context extractor based on the bounding box prediction; and (2) decoding target information from the context prompt with a context decoder. This approach extracts and updates context prompts discretely without explicit cross-frame correlation and highly depends on the accuracy of predictions, which may result in error accumulation and insufficient modeling of the target varying patterns. Furthermore, in vision-language tasks, there are multimodal references, yet most methods focus only on updating visual references, lacking an effective approach for jointly updating language and visual information.

Recently, the state space models, advanced by the LSSL~\cite{gu2021combining}, S4~\cite{gu2021efficiently}, GSS~\cite{mehta2022long}, and S4D~\cite{gu2022parameterization} have demonstrated exceptional performance in long-sequence modeling. Particularly, Mamba~\cite{gu2023mamba}, which uses selective variables to autoregressively model the sequential evolving neural states, has been noticed as a compelling alternative to Transformers on large-scale data. Yet, the utilization of state space for temporal multimodal feature modeling and updating is still under-investigated.

To jointly retain temporal information and update reference features adaptively, we explore the evolving process of Mamba's state space, in where MambaVLT memorizes long-term historical target features, and by which the model selectively updates the reference features. As shown in Figure \ref{fig:pipline}(b), our model captures temporal information through a continuously evolving state-space memory. Since Mamba autoregressively processes input sequences with the state space, the final state space in each layer inherently contains global features. Based on this observation, we design a state space memory and a state space evolving strategy to retain long-term multimodal information throughout the whole video. The evolved state space will be utilized to update the reference features adaptively. Compared with the method in Figure \ref{fig:pipline}(a), this approach not only enables context-aware multimodal modeling and reference feature updating but also provides a more elegant solution without extra network components.

Overall, MambaVLT introduces a time-evolving multimodal fusion module, which integrates a Hybrid Multimodal State Space (HMSS) block and a Selective Locality Enhancement (SLE) block. Each HMSS block consists of the temporal state space evolving mechanism and a modality-guided bidirectional scan. The temporal state space will transmit historical target features as shown in Figure \ref{fig:pipline}(b), while the modality-guided scan can dynamically update and fuse various modality features through different scan orders. After the HMSS module performs global modeling of cross-frame information, the SLE module will enhance the intra-modal dependency and inter-modal correlation of the current tracking frame through global receptiveness. Afterward, we present a modality-selection module to dynamically weigh the different modality reference features for search region feature refining to distinguish the reliability of different references at various time stamps.

Moreover, to analyze the effectiveness of state space memory and its ability to memorize long-term target information, we design a new tracking paradigm called semi-reference-free tracking, which aims to track without reference data input from the second search image in a video. To conclude, the main contributions of our work are:

\begin{itemize}
    \item We introduce MambaVLT, the first Mamba-based vision-language tracker, which is able to exploit the temporal information and update the reference features effectively and efficiently.
    % 合并原先2 3条
    \item We present a time-evolving multimodal fusion module that not only memorizes long-term target information for cross-frame information modeling and reference feature updating but also enhances the internal multimodal correlation of the current tracking frame.
    \item We conduct extensive experiments on the TNL2K~\cite{wang2021towards}, LaSOT~\cite{fan2019lasot}, OTB99~\cite{li2017tracking}, and MGIT~\cite{hu2024multi} benchmarks, which demonstrate the effectiveness of our MambaVLT.
\end{itemize}

\section{Related Work}
% related work写的太多 后面需要删减
% 1. TNL的进展

\subsection{Vision-language Tracking}
In the vision-language tracking task, there are three different reference settings: only target bounding box, only natural language, or both of them. Visual reference can provide more direct guidance, while natural language description can reveal details about the appearance and changes of the object over time. 

\vspace{1mm}
\noindent\textbf{Tracking by Initial Bounding Box} aims to continuously track a target throughout a video sequence based on the initial bounding box provided in the first frame. Siamese-based trackers~\cite{bertinetto2016fully,xu2020siamfc++,li2018high,li2019siamrpn++} utilize Siamese networks to extract visual features and locate targets by a matching module. To learn the historical changes of the target, Some trackers~\cite{fu2021stmtrack,sun2020fast,henriques2014high,nam2016learning,danelljan2017eco,yang2018learning,bhat2019learning} use previous prediction resutls for template update. TransT~\cite{chen2021transformer} introduces the Transformer architecture for visual tracking and achieves promising results. In addition, OSTrack~\cite{ye2022joint} and MixFormer~\cite{cui2022mixformer} construct simplified one-stream tracking pipelines with superior performance. However, tracking based on purely visual inference may lead to ambiguity in object identification.

\vspace{1mm}
\noindent\textbf{Tracking by Natural Language Specification} presents a unique approach with a more natural human-computer interaction way, which specifies the target with purely natural language description. Li \textit{et al}~\cite{li2017tracking} first defines this task and validates its effectiveness. They propose a paradigm conducting TNL task with separate grounding and tracking models which was adopted by following works~\cite{yang2020grounding,wang2021towards,li2022cross}. JointNLT~\cite{zhou2023joint} proposes a Transformer-based framework that unifies visual grounding and tracking and outperforms state-of-the-art algorithms. QueryNLT~\cite{shao2024context} then presents a context-aware fusion of visual and language references through query interactions to address the misalignment between language and visual information.
 
\vspace{1mm}
\noindent\textbf{Tracking by Language and Box Specification} specifies the target with a bounding box and natural language description. The work of Li \textit{et al}~\cite{li2017tracking} demonstrates that combining natural language description and initial target bounding box can enhance the tracking performance. SNLT~\cite{feng2021siamese} and VLT~\cite{guo2022divert} utilize natural language as an extra enhancement to aid visual features for tracking. Besides, both JointNLT and QueryNLT demonstrate great performance in this task. Recently, UVLTrack~\cite{ma2024unifying} proposes a unified Transformer-based architecture that can simultaneously model the above three reference settings. However, due to the inherent computation mechanisms of CNN and Transformer, the aforementioned methods struggle to learn long-range temporal information.

\subsection{State Space Models}
State space models (SSMs)~\cite{gu2021efficiently,fu2022hungry,smith2022simplified,gu2023mamba} have gained much attention because of their promising potential in long sequence modeling. Initially, the Structured State Space Sequence Model (S4)~\cite{gu2021efficiently} was proposed to model long-range dependencies in linear complexity. Based on S4, subsequent works including S5~\cite{smith2022simplified}, H3~\cite{fu2022hungry}, and Mamba~\cite{gu2023mamba} were proposed to improve the ability and efficiency of the model. Especially, Mamba outperforms Transformers in several long sequence NLP tasks with linear scalability due to its data-dependent selective state space mechanism and hardware implementation.

% a series of outstanding works后面需要添加引用
For the strong potential of Mamba in long sequence modeling, a series of outstanding works~\cite{vim,liu2024vmamba,he2024mambaad,weng2024mamballie,shi2024multi,zhang2024vfimamba} have been proposed in the visual domain. Vim~\cite{vim} and Vmamba~\cite{liu2024vmamba} adapt Mamba to visual classification tasks and released reliable pretrained models. They employ multidirectional scans to model the visual data. For video modeling, VideoMamba~\cite{li2024videomamba} applies S6 by concatenating 1D image sequences in temporal order. MambaIR~\cite{guo2024mambair} is the first to transfer the S6 model to the image restoration field. MTMamba~\cite{lin2024mtmamba} designed a Mamba-based dual-stream architecture for multi-task learning. CoupledMamba~\cite{li2024coupled} conducts multimodal fusion with coupling state chains of different modalities. 

Leveraging the autoregressive computation manner of Mamba, we propose a time-evolving mechanism to retain long-term target information, based on which a time-evolving multimodal fusion module is introduced to adaptively update reference features in the tracking. Meanwhile, a modality-selection module is designed to weigh the vision-language features for search region feature refining.

\section{MambaVLT}

\begin{figure*}[ht]
    \centering
    \includegraphics[width=1\linewidth]{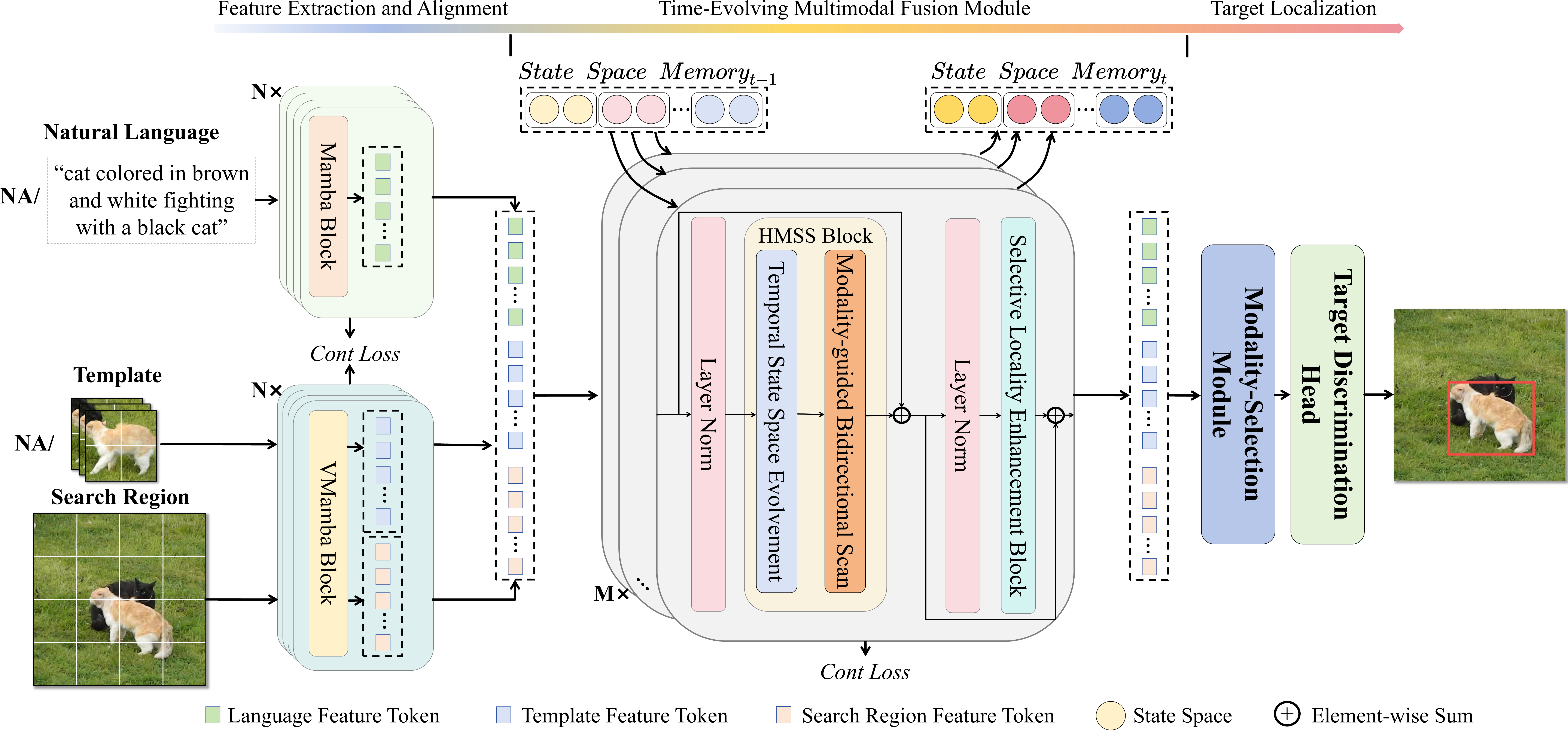}
    \caption{Overview of the MambaVLT. Given various modality reference settings, features are initially extracted and aligned, then forwarded to the time-evolving multimodal fusion module. Subsequently, these features are input into the localization module to obtain precise localization information. MambaVLT performs temporal information-aware vision-language tracking with adaptive reference feature updating. Note that 'NA' indicates when the corresponding reference is not provided.}
    \label{fig:backbone}
\end{figure*}

% 首先简述总体架构的流程和包含的内容
\subsection{Preliminaries: SSM and Mamba}
\textbf{State Space Model (SSM).} SSM is a continuous system that maps input sequence $x(t) \in \mathbb{R}$ to output sequence $y(t) \in \mathbb{R}$ with hidden state space $h(t) \in \mathbb{R}^{N}$, which can be formulated as follows:
\begin{equation}
\label{pre:ssm}
h^{\prime}(t)=\mathbf{A} h(t)+\mathbf{B} x(t), \quad y(t)=\mathbf{C} h(t) .
\end{equation}
where $\mathbf{A}$, $\mathbf{B}$, $\mathbf{C}$ are state transition matrices. The discrete counterpart,~\ie, discrete SSM, utilizes zero-order hold discretization with a timescale parameter $\boldsymbol{\Delta}$ to transform continuous parameters $\mathbf{A}$ and $\mathbf{B}$ into discrete parameters $\overline{\mathbf{A}}$ and $\overline{\mathbf{B}}$:
\begin{equation}
\label{pre:dssm}
\begin{aligned}
& \overline{\mathbf{A}}=\exp (\boldsymbol{\Delta} \mathbf{A}), \\
& \overline{\mathbf{B}}=(\boldsymbol{\Delta} \mathbf{A})^{-1}(\exp (\boldsymbol{\Delta} \mathbf{A})-\mathbf{I}) \cdot \boldsymbol{\Delta} \mathbf{B} .
\end{aligned}
\end{equation}
\textbf{Selective State Space Model (Mamba).} The above SSMs are still static systems for various inputs because of their data-independent parameters, which limits their ability to dynamically model sequences. To this end, Mamba generates $\mathbf{A}_i, \mathbf{B}_i, \boldsymbol{\Delta}_i$ based on the $i^{th}$ input $x_i$. The selective state space model can be written as:
\begin{equation}
\label{pre:mamba}
\begin{aligned}
& \overline{\bar{A}}_{i}=\exp \left(\Delta_{i} \boldsymbol{A}\right), \\
& \overline{\boldsymbol{B}}_{i}=\Delta_{i} \boldsymbol{B}_{i}, \\
& \boldsymbol{h}_{i} =\overline{\boldsymbol{A}}_{i} \boldsymbol{h}_{i-1}+\overline{\boldsymbol{B}}_{i} x_{i}, \\
& y_{i} =\boldsymbol{C}_{i} \boldsymbol{h}_{i}+\boldsymbol{D} x_{i}.
\end{aligned}
\end{equation}

\subsection{Overall Framework}

As shown in Figure \ref{fig:backbone}, the proposed MambaVLT is capable of jointly modeling different modality reference settings including the initial bounding box, natural language, or both. 
Firstly, we utilize a separate vision and language encoder for preliminary feature extraction. The input language description $l$ will be projected to language feature $F_l \in \mathbb{R}^{N_l \times C}$ with pretrained Mamba-based text encoder~\cite{gu2023mamba}. 
In particular, we use a template video clip to capture the appearance changes of the target explicitly. 
For the template video clip $z \in \mathbb{R}^{L \times 3 \times H_z \times W_z}$ and search region $x \in \mathbb{R}^{3 \times H_x \times W_x}$, they will be processed by shared Vmamba-based visual encoder~\cite{liu2024vmamba} to obtain template feature $F_z \in \mathbb{R}^{N_z \times \mathrm{C_v}}$ and serch region feature $F_x \in \mathbb{R}^{N_x \times \mathrm{C_v}}$. $L$ is the number of frames in the template video clip. Then, the language, template, and search region features will be concatenated to a unified 1D sequence $G$ for the time-evolving multimodal fusion, which will capture historical target information for reference features update and unified multimodal modeling.

% Target localization
To distinguish the reliability of visual and language references, the modality-selection module weighs and fuses multimodal references to refine the search region feature. Finally, the target discrimination head fully exploits the target and background information embedded in the search region feature to locate the target accurately. Additionally, the prediction head calculates a confidence score for each prediction to update the template video clip.

Moreover, we propose an intra-video and inter-video multimodal contrastive loss to align multimodal features during the feature fusion stages.
We firstly extract reference token $T$ with mean pooling operation based on reference features to calculate token-wise similarity $s_i$ with positive samples and negative samples to enhance the discriminative ability of features:
\begin{equation}
\label{method:sim}
s^i=\frac{\mathbf{T}\left(f^i\right)^{\top}}{\left\|\mathbf{T}\right\|_{2}\left\|f^i\right\|_{2}}.
\end{equation}
For intra-video contrastive loss $\mathcal{L}_{w}$, the positive sample is the target center token of the search region feature, and negative samples are $N_{w}^{n}$ most similar tokens in the search region background. For inter-video contrastive loss $\mathcal{L}_{o}$, the positive sample is the same as intra-video loss, while the negative samples are $N_{o}^{n}$ target center tokens from search regions of other video sequences.

\subsection{Time-Evolving Multimodal Fusion Module} 
\begin{figure*}[ht]
    \centering
    \includegraphics[width=1\linewidth]{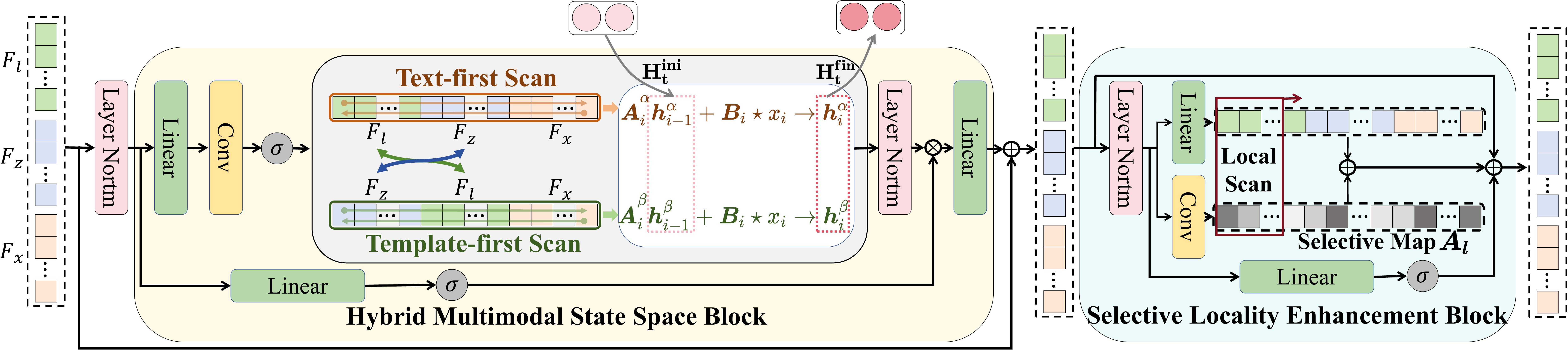}
    \caption{Overall pipeline of the Hybrid Multimodal State Space Block. The multimodal feature includes language feature $F_l$, template feature $F_z$ and search region feature $F_X$. The Hybrid Multimodal State Space block is for time-evolving global modeling and reference feature updating. Then, the Selective Locality Enhancement block will enhance the features of the current tracking frame. $\mathbf{{H}}^{ini}_{t}$ and $\mathbf{{H}}^{fin}_{t}$ denote the initial state space and final state space. local scan represents the linear attention scan. $\boldsymbol{A_l}$ represents the global selective map. }
    \label{fig:hmss}
\end{figure*}

Temporal information is crucial for dynamically adapting to target variations in vision-language tracking. Previous Transformer-based models mainly retain context information in a discrete manner. For continuous long-term target feature retention, MambaVLT presents a Time-evolving Multimodal Fusion (TEMF) module by harnessing the potential of the state space to enable unified feature modeling and adaptive reference information updating.

The TEMF module mainly consists of a Hybrid Multimodal State Space (HMSS) block and a Selective Locality Enhancement (SLE) block. Given a unified multimodal sequence $G$, the HMSS block first captures long-term temporal information by the time-evolving state space, based on which it models multimodal features and updates target reference information by a modality-guided bidirectional scan. After global cross-frame feature modeling, the SLE module performs a sliding window scan with a selective map $A_l$ to enhance multimodal features of the current time stamp through a global receptiveness. Formally,
\begin{equation}
\label{method:TEMF}
G^{\prime}=\boldsymbol{\phi_{SLE}}\left(\left(\boldsymbol{\phi_{HMSS}}\left(G\right)\right)\right).
\end{equation}

\vspace{1mm}
\noindent\textbf{Hybrid Multimodal State Space Block.} As shown in Figure \ref{fig:hmss}, the Hybrid Multimodal State Space (HMSS) block integrates the temporal hybrid state space evolving mechanism for temporal information retention and a modality-guided directional scan for dynamic features update.

In the temporal hybrid state space evolving mechanism, we construct a multi-level state space memory $SS =\{\{\mathbf{{H}}^{{fin}_i, \alpha}_{t-1},\enspace \mathbf{{H}}^{{fin}_i, \beta}_{t-1}\}|i \in {1,2,...,M} \}$ to store different final state space $\mathbf{{H}}^{fin}$ of TEMF modules, where $ \alpha$ and $ \beta$ denote text-first and template-first scan. $M$ is the number of TEMF modules. Since Mamba processes sequences with the state space autoregressively, the final state space will inherently contain global information of processed tokens. As the state space memory processes each frame of the video sequence and updates the template video clip, the state space memory evolves temporally and memorizes long-term target information naturally. At the prior of each HMSS module, we derive the initial state space from the state space memory and a learnable state space $\mathbf{{H}}^{l}$:
\begin{equation}
\label{method:stateevo}
\mathbf{{H}}^{ini}_{t}= a \mathbf{{H}}^{l} + (1-a) \mathbf{{H}}^{fin}_{t-1}.
\end{equation}
where $\mathbf{{H}}^{fin}_{t-1}$ represents the final state space of the last time stamp in state space memory. $a$ is the trade-off parameter.

Captured by the temporal target information by multi-level state space memory, the HMSS block will perform a modality-guided bidirectional scan to adaptively update reference features and fuse multimodal information, which is based on the prior insight that different scan orders in Mamba will influence the modality feature. As shown in Figure \ref{fig:hmss}, the HMSS block will conduct bidirectional scans based on text-first order $\alpha$ and template-first order $\beta$, in which search region feature will always be placed at the end of the sequence to gather reference information. By changing the order of the text and template features, they will serve as guiding information to direct the update and fusion of the features respectively. Different from previous multi-direction scan methods~\cite{vim, li2024videomamba} which use completely different parameters for multidirectional scan, HMSS block mainly utilizes shared parameters including $\overline{\boldsymbol{B}}, \boldsymbol{C}$ and $\boldsymbol{D}$ to reduce parameter redundancy and model the overall perception of target information. We use distinct parameters $\overline{\boldsymbol{A}}$ as state space update gates in different scan orders to adaptively update reference features and model search region features. This process can be formulated as:
\begin{equation}
\label{method:mgb}
\begin{aligned}
& \boldsymbol{h}_{i}^{\alpha} =\overline{\boldsymbol{A}}_{i}^{\alpha} \boldsymbol{h}_{i-1}^{\alpha}+\overline{\boldsymbol{B}}_{i} \star x_{i}, \\
& \boldsymbol{h}_{i}^{\beta} =\overline{\boldsymbol{A}}_{i}^{\beta} \boldsymbol{h}_{i-1}^{\beta}+\overline{\boldsymbol{B}}_{i} \star x_{i}, \\
& y_{i} =(\boldsymbol{C}_{i} \star \boldsymbol{h}_{i}^{\alpha} + \boldsymbol{C}_{i} \star \boldsymbol{h}_{i}^{\beta})/2 +\boldsymbol{D} \star x_{i}.
\end{aligned}
\end{equation}

The $\overline{\boldsymbol{B}}, \boldsymbol{C}$ and $\boldsymbol{D}$ are control gates for overall feature extraction. $\overline{\boldsymbol{A}}_{i}^{\alpha}$ and $\overline{\boldsymbol{A}}_{i}^{\beta}$ are utilized for language-guided $\alpha$ and template-guided $\beta$ autoregressive feature update. $\star$ denotes the Hadamard product performed after aligning the parameters and features according to the modality order.

\vspace{1mm}
\noindent \textbf{Selective Locality Enhancement Block.} After HMSS performs global modeling of multimodal information on cross-frame temporal dimension, we introduce a Selective Locality Enhancement (SLE) block to enhance features of the current time stamp. Linear attention is known for reducing the computation cost of softmax attention to O(N). Inspired by previous linear attention work~\cite{beltagy2020longformer,han2024demystify} and the carefully designed architecture of Mamba, we propose a 1D local scan method with global selective receptiveness. In classic linear attention, only top layers have access to nearly global representation~\cite{beltagy2020longformer} which is critical to multimodal correlation modeling. To address this issue, as shown in the right part of Figure \ref{fig:hmss}, we introduce a global selective map $\boldsymbol{A}_{l}$ to enhance the global perception capability of the SLE block. $\boldsymbol{A}_{l}$ is obtained by performing a convolution operation on the output of the HMSS block to extract the inherent global selective information in the HMSS block. Then the sequence added with a global selective map will be enhanced by linear attention scan $\boldsymbol{\gamma}$. Moreover, the SLE block employs several Mamba-like control gates including $\boldsymbol{B}_{l}$ and $\boldsymbol{D}_{l}$ to process the sequence. The SLE block can be formulated as:
\begin{equation}
\label{method:la}
\begin{aligned}
& \boldsymbol{h}_{t} =\boldsymbol{A}_{l}+\boldsymbol{B}_{l} G, \\
& G^{\prime} =\boldsymbol{\gamma}(\boldsymbol{h}_{l})+ \boldsymbol{D}_{l} G.
\end{aligned}
\end{equation}
$\boldsymbol{B}_{l}, \boldsymbol{D}_{l}$ denote the input gate and residual gate. $\boldsymbol{\gamma}$ represents the sliding window linear attention scan to enhance the intra-modal dependency. $\boldsymbol{A}_{l}$ is responsible for extracting the selective information as a global map. In this manner, the SLE block can selectively enhance different reference features and search region features while maintaining linear computational complexity.

\subsection{Modality-Selection Module}
\label{method:ms}
\begin{figure}[t]
    \centering
    \includegraphics[width=0.9\linewidth]{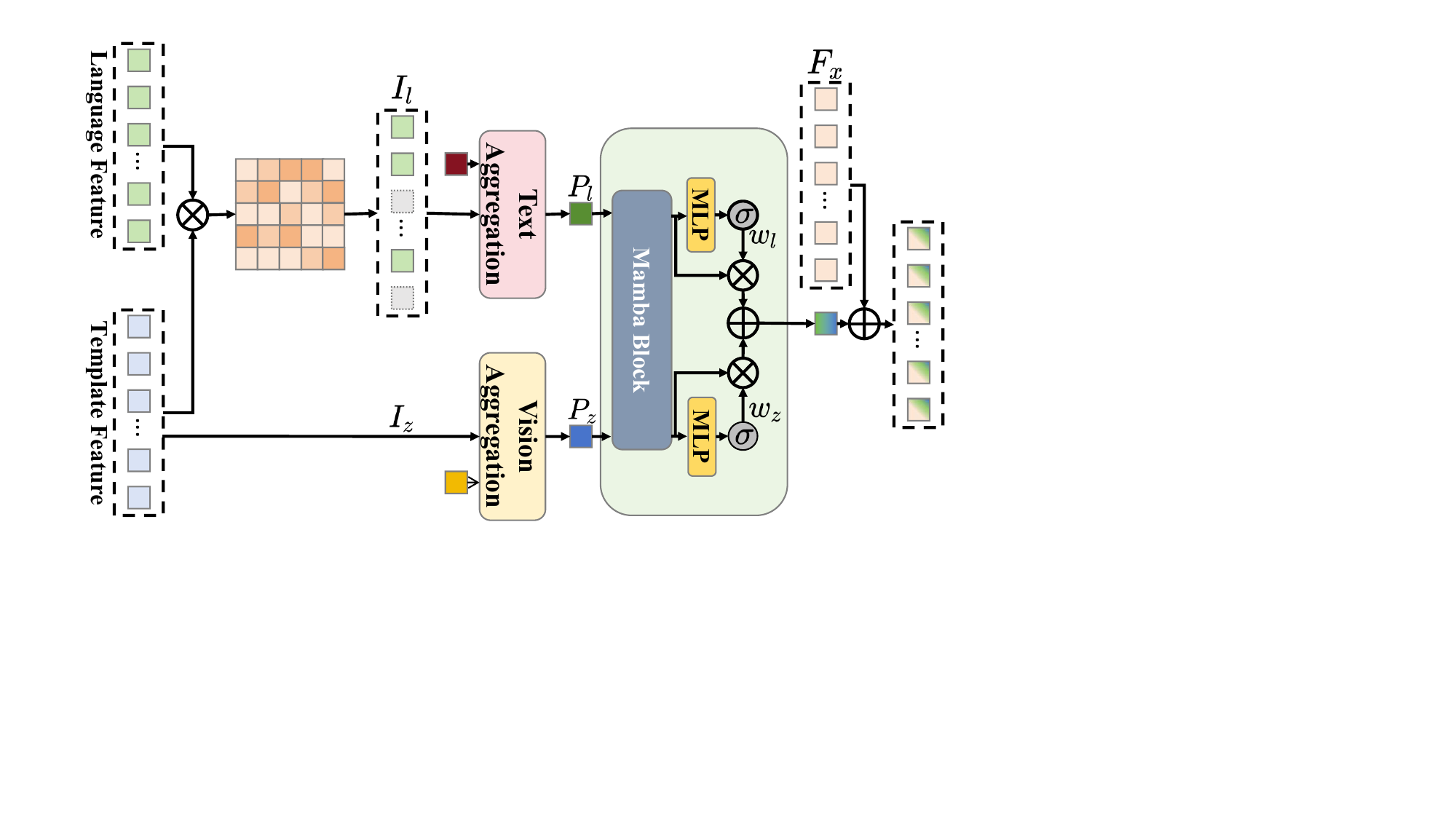}
    \caption{Overview of modality-selection module. $w_l$ and $w_z$ represents the weights of language invariant clue $P_l$ and template invariant clue $P_z$.}
    \label{fig:ms}
\end{figure}
\par In the time-evolving multimodal fusion module, the search region feature dynamically interacts with multimodal temporal features in a different fixed order. However, in different tracking frames, the reliability of the language and template features may vary due to the target motion and appearance changes. Therefore, we further employ a Modality-selection module to selectively fuse multimodal reference features for search region feature refining. As shown in Figure \ref{fig:ms}, it will firstly extract invariant language information $I_l$ and template information $I_z$. Because the template feature can naturally reflect the invariant target appearance feature, we compute the similarity between the language and template features, based on which we extract N language tokens with the highest visual similarity as the final invariant language information.

Subsequently, language and vision query decoders are introduced to aggregate the invariant language and vision target clue: $P_l$ and $P_z$. A Mamba-based selective block is then employed to weigh the $P_l$ and $P_z$ for language and vision clues fusion. The selected invariant reference clue will be used to refine the search region feature for more accurate target localization.

\subsection{Training Objective}
\label{method:loss}
\par The contrastive loss is consist of intra-video contrastive loss $\mathcal{L}_{c_w}$ and inter-video contrastive loss $\mathcal{L}_{c_o}$. Formally,
\begin{equation}
\label{method:contrastive}
\mathcal{L}_{c}^{i}=-\log \left(\frac{e^{s_{c}^{p}}}{e^{s_{c}^{p}}+\sum_{k=1}^{N_{c}^{n}} e^{s_{c}^{n_k}}}\right) .
\end{equation}

Token-wise similarity $s$ is calculated based on Equation \ref{method:sim}. A binary cross-entropy loss is employed for target score map loss $\mathcal{L}_{tgt}$, whose groundtruth is generated based on the bounding box. We utilize the same training objectives of center score map $\mathcal{L}_{cls}$ and bounding box loss $\mathcal{L}_{bbox}=\lambda_{1} \mathcal{L}_{1}+\lambda_{giou} \mathcal{L}_{giou}$ as OSTrack~\cite{ye2022joint}. $\lambda$ denotes different loss weights. The whole training objectives can be summarized as:
\begin{equation}
\label{method:allloss}
\mathcal{L}=\lambda_{bbox}\mathcal{L}_{bbox}+\lambda_{tgt}\mathcal{L}_{tgt}+\lambda_{cls}\mathcal{L}_{cls}+\lambda_{c_w}\mathcal{L}_{c_w}+\lambda_{c_o}\mathcal{L}_{c_o}.
\end{equation}
\section{Experiments}

\begin{table*}[ht]
\centering
\small
\setlength{\extrarowheight}{-5.5pt}
\addtolength{\extrarowheight}{\aboverulesep}
\addtolength{\extrarowheight}{\belowrulesep}
\setlength{\aboverulesep}{0pt}
\setlength{\belowrulesep}{0pt}
\caption{Comparison of our method with state-of-the-art approaches on TNL2k, LaSOT and OTB99 datasets. The best and second-best results are highlighted in \textcolor{red}{red} and \textcolor{blue}{blue} respectively.}
\label{table:expthree}
\begin{tabular}{c|c|ccc|ccc|ccc} 
\toprule
\multirow{2}{*}{\textbf{Tracker}}      & \multirow{2}{*}{\textbf{Reference}} & \multicolumn{3}{c|}{\textbf{TNL2K }}           & \multicolumn{3}{c|}{\textbf{LaSOT }}           & \multicolumn{3}{c}{\textbf{OTB99 }}             \\ 
\cline{3-11}
                                       &                                     & \textit{AUC} & \textit{Prec} & \textit{N Prec} & \textit{AUC} & \textit{Prec} & \textit{N Prec} & \textit{AUC} & \textit{Prec} & \textit{N Prec}  \\ 
\hline
SiamRPN++~\cite{li2019siamrpn++}       & BBOX                                  & 41.3         & 41.2          & 48.0            & 49.6         & 49.1          & 56.9            & -            & -             & -                \\
AutoMatch~\cite{zhang2021learn}        & BBOX                                & 47.2         & 43.5          & -               & 58.3         & 59.9          & 67.4            & -            & -             & -                \\
TriDiMP~\cite{wang2021transformer}     & BBOX                                 & 52.3         & 52.8          & -               & 63.9         & 61.4          & -               & -            & -             & -                \\
TransT~\cite{chen2021transformer}      & BBOX                                & 50.7         & 51.7          & -               & 64.9         & 69.0          & 73.8            & -            & -             & -                \\
SwinTrack-B~\cite{lin2022swintrack}   & BBOX                                & -            & 55.9          & \textcolor{blue}{57.1}            & 61.3         & 76.5          & -               & -            & -             & -                \\
OSTrack-256~\cite{ye2022joint}        & BBOX                                 & 54.3         & -             & -               & 69.1         & 75.2          & \textcolor{blue}{78.7}            & -            & -             & -                \\
GRM~\cite{gao2023generalized}         & BBOX                                & -            & -             & -               & \textcolor{red}{69.9}         & \textcolor{red}{75.8}          & \textcolor{red}{79.3}            & -            & -             & -                \\
UVLTrack-B~\cite{ma2024unifying}      & BBOX                                & \textcolor{blue}{62.7}         & \textcolor{blue}{65.4}          & -               & \textcolor{blue}{69.4}         & \textcolor{blue}{74.9}          & -               & \textcolor{blue}{69.3}         & \textcolor{blue}{90.1}          & \textcolor{blue}{84.3}             \\
\rowcolor[rgb]{0.831,0.831,0.831} Ours & BBOX                                & \textcolor{red}{63.3}         & \textcolor{red}{65.8}          & \textcolor{red}{87.5}            & 65.0         & 69.5          & 76.6            & \textcolor{red}{71.6}         & \textcolor{red}{92.9}          & \textcolor{red}{87.4}            \\ 
\hline
TNLS-II~\cite{li2017tracking}        & NL                                  & -            & -             & -               & -            & -             & -               & 25.0         & 29.0          & -                \\
RVTNLN~\cite{feng2019robust}         & NL                                  & -            & -             & -               & -            & -             & -               & 54.0         & 56.0          & -                \\
RTTNLD~\cite{feng2020real}           & NL                                  & -            & -             & -               & 28.0         & 28.0          & -               & 54.0         & 78.0          & -                \\
GTI~\cite{yang2020grounding}         & NL                                  & -            & -             & -               & 47.8         & 47.6          & -               & 58.1         & 73.2          & -                \\
TNL2K-1~\cite{wang2021towards}       & NL                                  & 11.4         & 6.4           & 11.0            & 51.1         & 49.3          & -               & 19.0         & 24.0          & -                \\
CTRNLT~\cite{li2022cross}            & NL                                  & 14.0         & 9.0           & -               & 52.0         & 51.0          & -               & 53.0         & 72.0          & -                \\
JointNLT~\cite{zhou2023joint}        & NL                                  & 54.6         & 55.0          & \textcolor{blue}{70.6}            & \textcolor{blue}{56.9}         & \textcolor{blue}{59.3}          & \textcolor{red}{64.5}            & 59.2         & 77.6          & -                \\
QueryNLT~\cite{shao2024context}      & NL                                  & 53.3         & 53.0          & 70.4            & 54.2         & 55.0          & 62.5            & \textcolor{red}{61.2}         & \textcolor{red}{81.0}          & \textcolor{red}{73.9}             \\
UVLTrack-B~\cite{ma2024unifying}     & NL                                  & \textcolor{blue}{55.7}         & \textcolor{blue}{57.2}          & -               & \textcolor{red}{57.2}         & \textcolor{red}{61.0}          & -               & \textcolor{blue}{60.1}         & 79.1          & -                \\
\rowcolor[rgb]{0.831,0.831,0.831} Ours & NL                                  & \textcolor{red}{58.4}         & \textcolor{red}{58.9}          & \textcolor{red}{80.9}            & 55.8         & 57.2         & \textcolor{blue}{63.7}           & 58.9         & \textcolor{blue}{79.2}          & \textcolor{blue}{72.0}             \\ 
\hline
TNLS-III~\cite{li2017tracking}      & NL$\&$BBOX                               & -            & -             & -               & -            & -             & -               & 55.0         & 72.0          & -                \\
RVTNLN~\cite{feng2019robust}        & NL$\&$BBOX                               & 25.0         & 27.0          & 34.0            & 50.0         & 56.0          & -               & 67.0         & 73.0          & -                \\
RTTNLD~\cite{feng2020real}          & NL$\&$BBOX                               & 25.0         & 27.0          & 33.0            & 35.0         & 35.0          & -               & 61.0         & 79.0          & -                \\
SNLT~\cite{feng2021siamese}         & NL$\&$BBOX                               & 27.6         & 41.9          & -               & 54.0         & 57.6          & -               & 66.6         & 80.4          & -                \\
TNL2K-2~\cite{wang2021towards}      & NL$\&$BBOX                               & 41.7         & 42.0          & 50.0            & 51.0         & 55.0            & -               & 68.0         & 88.0          & -                \\
JointNLT~\cite{zhou2023joint}       & NL$\&$BBOX                               & 56.9         & 58.1          & 73.6            & 60.4         & 63.6          & 69.4            & 65.3         & 85.6          & 79.5             \\
QueryNLT~\cite{shao2024context}     & NL$\&$BBOX                               & 57.8         & 58.7          & \textcolor{blue}{75.6}            & 59.9         & 63.5          & \textcolor{blue}{69.6}            & 66.7         & 88.2          & \textcolor{blue}{82.4}             \\
UVLTrack-B~\cite{ma2024unifying}    & NL$\&$BBOX                               & \textcolor{blue}{63.1}         & \textcolor{blue}{66.7}          & -               & \textcolor{red}{69.4}         & \textcolor{red}{75.9}          & -               & \textcolor{blue}{69.3}         & \textcolor{blue}{89.9}          & -                \\
\rowcolor[rgb]{0.831,0.831,0.831} Ours & NL$\&$BBOX                               & \textcolor{red}{66.5}         & \textcolor{red}{69.9}         & \textcolor{red}{90.9}            & \textcolor{blue}{66.6}         & \textcolor{blue}{71.0}          & \textcolor{red}{77.3}            & \textcolor{red}{72.2}         & \textcolor{red}{94.4}          & \textcolor{red}{88.1}             \\
\bottomrule
\end{tabular}
\end{table*}

\subsection{Implementation Details}
\textbf{Network Configuration.} For vision inputs, the template and search region size are set to be $128 \times 128$ and $256 \times 256$. In the grounding task, the search region is resized such that its long edge is equal to 256. For the tracking task, the template is cropped from the first image and the scale factor is 2. The search region is cropped based on the last prediction bounding box and the scale factor is 4. For language inputs, the maximum length is 40. We utilize the first 4 layers of Mamba-130m~\cite{gu2023mamba} as text encoder and the 4-stage Vmamba-tiny~\cite{liu2024vmamba} as visual encoder. Notably, we modify the downsample layer of Vmamba to set the image patch size to 16. Also, we construct 4 time-evolving multimodal fusion modules. 
\vspace{1mm}
\par \noindent \textbf{Training Details.} We utilize the official training splits of OTB99~\cite{li2017tracking}, LaSOT~\cite{fan2019lasot}, TNL2K~\cite{wang2021towards}, MGIT~\cite{hu2024multi}, RefCOCOg-google~\cite{mao2016generation}, and GOT-10k~\cite{huang2019got} to train our model. We use Adam to optimize the model and the learning rate is 0.0005. The weight decay coefficient is 0.05. We train our model for 300 epochs. The batch size is 8. We utilize common data augmentation methods including horizontal flip, translation, and color jittering.

\begin{table}
\centering
\small
\setlength{\extrarowheight}{-5.5pt}
\addtolength{\extrarowheight}{\aboverulesep}
\addtolength{\extrarowheight}{\belowrulesep}
\setlength{\aboverulesep}{0pt}
\setlength{\belowrulesep}{0pt}
\caption{Comparison of our method with the latest approaches on the MGIT dataset based on the official reproduction results.}
\label{table:expmgit}
\begin{tabular}{c|c|ccc} 
\toprule
\multirow{2}{*}{\textbf{Tracker}}      & \multirow{2}{*}{\textbf{Reference}} & \multicolumn{3}{c}{\textbf{MGIT}}               \\ 
\cline{3-5}
                                       &                                     & \textit{AUC} & \textit{Prec} & \textit{N Prec}  \\ 
\hline
PriDiMP~\cite{wang2021transformer}     & BBOX                                  & -         & 29.6          & 60.2             \\
TransT~\cite{chen2021transformer}      & BBOX                                  & -         & 44.7          & 67.0             \\
OSTrack~\cite{ye2022joint}         & BBOX                                  & -         & 47.6          & 70.6             \\
GRM~\cite{gao2023generalized}          & BBOX                                  & -         & \textcolor{blue}{50.0}          & \textcolor{blue}{71.8}             \\
\rowcolor[rgb]{0.831,0.831,0.831} Ours & BBOX                                  & \textcolor{red}{65.7}         & \textcolor{red}{51.6}          & \textcolor{red}{72.9}             \\ 
\hline
\rowcolor[rgb]{0.831,0.831,0.831} Ours & NL                                  & \textcolor{red}{64.6}         & \textcolor{red}{50.3}          & \textcolor{red}{71.2}             \\ 
\hline
SNLT ~\cite{feng2021siamese}           & NL$\&$BBOX                                & -          & 0.4           & 22.6             \\
VLT\_SCAR~\cite{guo2022divert}         & NL$\&$BBOX                                & -         & 11.6          & 35.4             \\
VLT\_TT~\cite{guo2022divert}           & NL$\&$BBOX                                & -         & 31.8          & 60.2             \\
JointNLT~\cite{zhou2023joint}          & NL$\&$BBOX                                & -         & \textcolor{blue}{44.5}          & \textcolor{red}{78.6}             \\
\rowcolor[rgb]{0.831,0.831,0.831} Ours & NL$\&$BBOX                                & \textcolor{red}{69.9}         & \textcolor{red}{58.9}          & \textcolor{blue}{78.0}             \\
\bottomrule
\end{tabular}
\end{table}

\subsection{The Analysis of State Space}
% 通过图可以看出，随着追踪的深入以及视角的不断变化，SRF任务不使用reference信息，反而效果超越UVLTrack，这证明了state space记忆的可靠性以及能够长期记忆目标的信息的能力。而在应对干扰项的时候，能够捕获长序列时序依赖的state space的追踪更加稳定，这也是state space的另外优点。

\par To analyze the effectiveness of state space memory and its capability to retain long-term target information, we design a new tracking paradigm called \textbf{semi-reference-free} (SRF) tracking. In the semi-reference-free tracking, the reference data (language or initial bounding box) is used by tracker \textbf{only} in the first frame. The tracker needs to extract and retain the target information embedded within the reference data and subsequently locate the target in search regions solely through the retained target information, without relying on reference data. 

\begin{figure}
    \centering
    \includegraphics[width=1\linewidth]{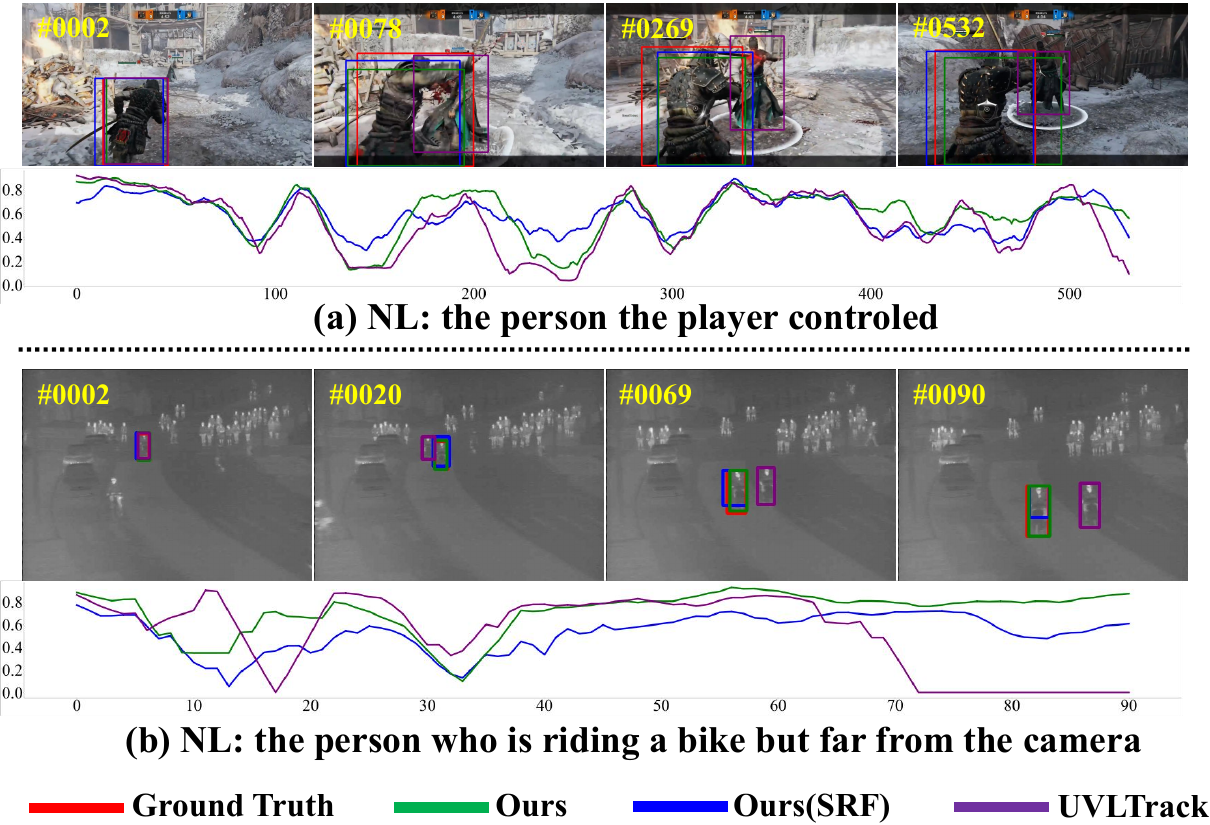}
    \caption{Qualitative comparison of NL$\&$BBOX tracking task on two challenging sequences to analyze the effectiveness of state space. The line graphs represent the IoU of different trackers for each frame. 
    The SRF means semi-reference-free tracking setting. 
    % The main challenge factors of (a) and (b) are target fast movement and distractors.
    }
    \label{fig:expss}
\end{figure}

As shown in Figure \ref{fig:expss}, in the NL$\&$BBOX tracking task, we conduct qualitative comparisons on two sequences of different trackers to analyze the ability of the proposed state space evolving mechanism. 
% The challenges with these two sequences are target fast movement and target loss, respectively.
The main challenges with these two sequences are target fast movement and distractors, respectively.
As shown in the line plot, MambaVLT with the semi-reference-free setting generally outperforms UVLTrack with normal settings, which demonstrates that the proposed state space memory can efficiently extract target information from references and retain long-term target information during the tracking process.
As shown in Figure \ref{fig:expss}(b), our model is capable of continuously tracking the target even with the semi-reference-free setting in a blurred environment with multiple distractors, which shows the selectivity of the state space memory can increase the discriminative ability of the model.

\subsection{Comparison with state-of-the-art trackers}

\begin{table}[ht]
\centering
\small
\setlength{\extrarowheight}{-5.5pt}
\addtolength{\extrarowheight}{\aboverulesep}
\addtolength{\extrarowheight}{\belowrulesep}
\setlength{\aboverulesep}{0pt}
\setlength{\belowrulesep}{0pt}
\caption{Analysis of different components in MambaVLT}
\label{table:ablation}
\begin{tabular}{ccccccc} 
\toprule
\multicolumn{1}{c}{\multirow{3}{*}{Variants}} & \multicolumn{6}{c}{TNL2k}                                                                \\ 
\cline{2-7}
\multicolumn{1}{c}{}                         & \multicolumn{2}{c}{BBOX}    & \multicolumn{2}{c}{NL}      & \multicolumn{2}{c}{NL\&BBOX}   \\ 
\cline{2-7}
\multicolumn{1}{c}{}                         & \textit{AUC} & \textit{Prec} & \textit{AUC} & \textit{Prec} & \textit{AUC} & \textit{Prec}  \\ 
\hline
baseline                                     & 60.9         & 62.7         & 55.3         & 55.0         & 62.6         & 65.1          \\
+THSS                                        & 62.1         & 64.2         & 56.8         & 57.6         & 64.5         & 67.3          \\
+MgB                                         & 62.5         & 64.3         & 57.3         & 57.9         & 65.3         & 68.2          \\
+MS                                          & 63.0         & 65.1         & 57.8         & 58.5         & 65.8         & 69.0          \\
\rowcolor[rgb]{0.831,0.831,0.831} +SLE       & 63.3         & 65.8         & 58.4         & 58.9         & 66.5         & 69.9          \\
\bottomrule
\end{tabular}
\end{table}
\textbf{Tracking by Initial Bounding Box (BBOX).} In this section, we compare our method with state-of-the-art trackers using only initial Bounding Box for target specification on four datasets including TNL2K~\cite{wang2021towards}, LaSOT~\cite{fan2019lasot}, OTB99~\cite{li2017tracking}, and MGIT~\cite{hu2024multi}. We utilize the Area Under the Curve (AUC) of the success plot and the tracking precision (Prec) as the main metrics to rank trackers. As shown in Table \ref{table:expthree} and \ref{table:expmgit}, our MambaVLT outperforms previous trackers in TNL2k, OTB99, and MGIT but performs less effectively on the LaSOT dataset. However, MambaVLT still achieves better results than some Transformer-based trackers including TransT and SwinTrack. MambaVLT outperforms the best trackers on the TNL2k and OTB99 datasets in terms of AUC by 0.6$\%$, and 2.3$\%$, respectively. It also surpasses the GRM~\cite{gao2023generalized} on the MGIT in terms of PRE by 1.6$\%$.
\vspace{1mm}
\par \noindent \textbf{Tracking by Language Specification (NL).} We conduct experiments in tracking by language specification task across the four aforementioned benchmarks with state-of-the-art trackers. Our method achieves optimal performance on the TNL2k and MGIT datasets. However, the suboptimal results on the LaSOT and OTB99 datasets may be due to their limited capability to track targets based on ambiguous textual descriptions.
\vspace{1mm}
\par \noindent \textbf{Tracking by Language and Bounding Box (NL$\&$BBOX).} We further evaluate MambaVLT in tracking by language and bounding box task with the latest trackers. The benchmarks are TNL2k, LaSOT, OTB99 and MGIT. MambaVLT shows more superior performance, achieving AUC improvements of 3.4\% and 2.9\% on the TNL2K and OTB99. Besides, it improves the PRE metric by 14.4\% on MGIT. In the LaSOT dataset, MambaVLT performs below UVLTrack but achieves a 6.7\% higher AUC than QueryNLT. Compared with previous Transformer-based trackers which utilize discrete context prompt, MambaVLT achieves better performance by introducing a continuous time-evolving state space memory to capture long-term multimodal temporal information.

\subsection{Ablation Study}
\begin{figure}[t]
    \centering
    \includegraphics[width=1\linewidth]{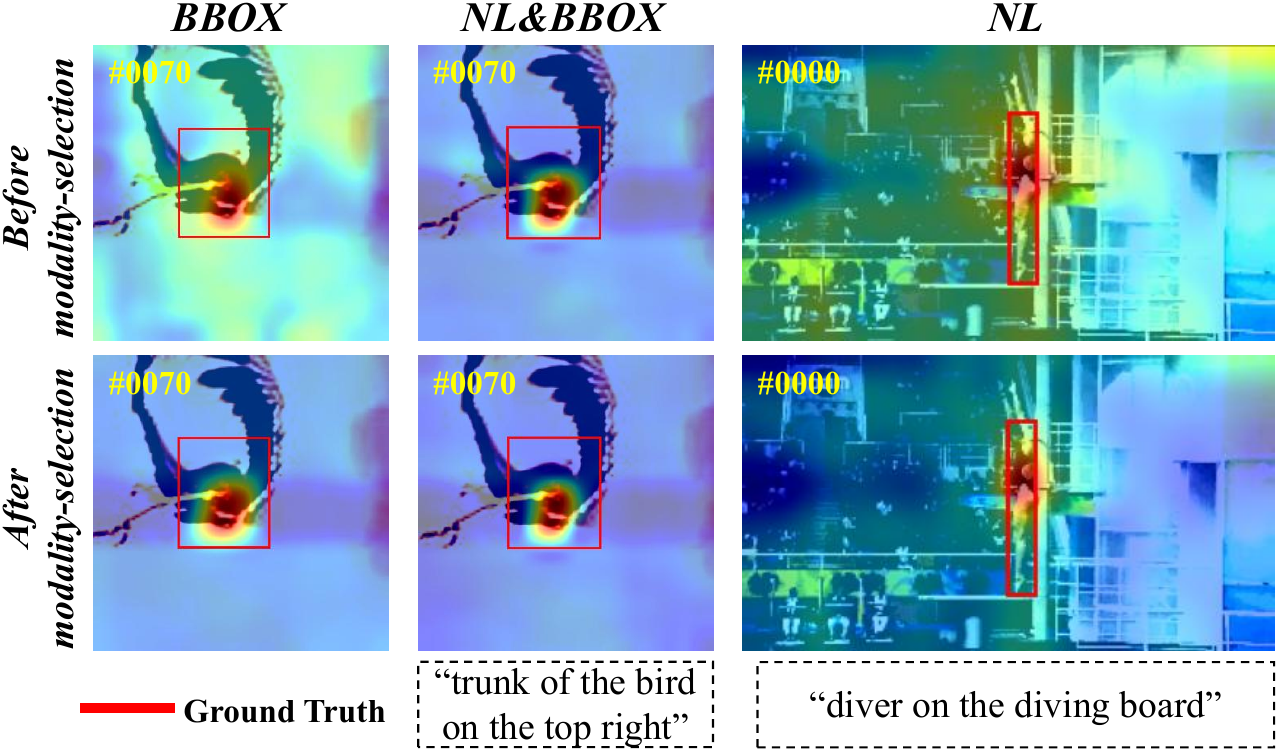}
    \caption{Visualization of the similarity between reference token and search region before and after the modality-selection module.}
    \label{fig:expsim}
\end{figure}

In this section, we analyze the effectiveness of different main components in MambaVLT, which includes five variants of our model as shown in Table \ref{table:ablation}. The baseline is MambaVLT without time-evolving hybrid state space (THSS), modality-guided bidirectional scan (MgB), modality-selection (MS) module and selective locality enhancement (SLE) block. The time-evolving hybrid state space brings 1.2\%, 1.5\% and 1.9\% AUC increase for BBOX, NL and NL\&BBOX tasks, respectively, which shows the great ability of our time-evolving state space to capture long-term temporal information to adapt to target changes in vision-language tracking.
Collaborating with time-evolving state space, the modality-guided scan bidirectional scan improves model performance by dynamically modeling and updating reference features through text-first and template-first scans. 

The introduction of the modality-selection module increases the ability of the model to weigh the importance of different modality information, further improving overall performance. 
Furthermore, Figure \ref{fig:expsim} shows the similarity between the reference token and search region feature before and after the modality-selection module, which indicates the refinement of the search region by the modality-selection module can enhance the discriminative ability of model in target localization. 
The performance increase after adding the selective locality enhancement block demonstrates that its capability to further enhance multimodal features in the current tracking frame.
\section{Conclusion}
In this work, we propose a Mamba-based vision-language tracking framework with a time-evolving state space to capture long-term continuous target information, based on which the proposed modality-guided bidirectional scan will model and update the multimodal features in a cross-frame manner. Besides, the selective locality enhancement block will enhance the features in the current tracking frame. Moreover, we present a modality-selection module to dynamically weigh the different modality reference features for search region feature refining. Our model achieves favorable performance against state-of-the-art algorithms on four vision-language tracking datasets.
{
    \small
    \bibliographystyle{ieeenat_fullname}
    \bibliography{main}
}

\renewcommand\thesection{\Alph{section}}
\renewcommand{\thefigure}{\Alph{figure}}
\renewcommand{\thetable}{\Alph{table}}
\setcounter{section}{0} % 重置章节计数器
\setcounter{figure}{0}  % 重置图片计数器
\setcounter{table}{0}   % 重置表格计数器
\clearpage
\setcounter{page}{1}
\maketitlesupplementary

\section{Additional Implementation Details}
\subsection{Target Discrimination Head}
We proposed the target discrimination head to exploit the discriminative information from the reference feature to locate the target. We will first extract a unified reference token $T_{uni}$. In the BBOX and NL tasks, $T_{uni}$ is extracted based on the template feature and the language feature, respectively. In the NL\&BBOX task, we perform mean pooling on the template and language features to obtain $T_{uni}$. Subsequently, we apply two separate linear layers to transform $T_{uni}$ into target token $T_{tgt}$ and background token $T_{bgd}$. $T_{tgt}$ and $T_{bgd}$ are used to compute the similarity with the search region feature, generating the target score and the background score for each search region token. These scores are then utilized to select the final output bounding box. We employ the binary cross-entropy target score map loss $\mathcal{L}_{tgt}$, whose groundtruth is generated based on the bounding box, as the contrastive learning loss for target discrimination. Finally, the prediction with a target score exceeding the threshold will be used to update the template video clip. The threshold is set to 0.8.

\subsection{Training Settings}
% 放对比学习细节，训练其他参数的细节
In the intra-video contrastive learning, we utilize 1 positive sample and 8 negative samples.  In the inter-video contrastive learning, we utilize 1 positive sample and 224 negative samples. The multimodal contrastive learning is performed in the last two layers of the preliminary feature extraction stage and every module of the time-evolving multimodal fusion module.

\section{More Experimental Results}
\subsection{Extensive Experiments on MGIT}
%放MGIT结果
In the MGIT~\cite{hu2024multi} dataset, in addition to the language descriptions of the targets in the first frames, it also provides corresponding natural language specifications of the targets in certain subsequent frames. Therefore, \textbf{without retraining} the model, we update the language information during inference with the latest natural language description, to evaluate whether the state space memory can update the target feature based on the new description, thereby improving tracking accuracy. Notably, all the experiments are conducted using the action granularity of the MGIT dataset.

\begin{table}
\centering
\small
\setlength{\extrarowheight}{-5.5pt}
\renewcommand{\arraystretch}{1.0}
% \addtolength{\extrarowheight}{\aboverulesep}
% \addtolength{\extrarowheight}{\belowrulesep}
% \setlength{\aboverulesep}{0pt}
% \setlength{\belowrulesep}{0pt}
\caption{Extensive experiments of natural language updating on MGIT. * denotes the results obtained by updating the language descriptions in the inference process without retraining the model.}
\label{sup:mgit}
\begin{tabular}{lcccc} 
\toprule
\multicolumn{1}{l}{\multirow{2}{*}{Tracker}} & \multicolumn{4}{c}{\makecell{MGIT}}                                                                                                                       \\ 
\cline{2-5}
\multicolumn{1}{c}{}                         & \makecell{\textit{AUC}} & \textit{Prec} & \textit{N prec} & \textit{${{SR}_{IoU}}$}  \\ 
\hline
\multicolumn{5}{c}{\makecell{BBOX}}                                                                                                                                                                       \\ 
\hline
\makecell[l]{MambaVLT}                                      & 65.7         & 51.6          & 72.9            & 60.4                          \\ 
\hline
\multicolumn{5}{c}{\makecell{NL}}                                                                                                                                                                         \\ 
\hline
\makecell[l]{MambaVLT}                                      & 64.6         & 50.3          & 71.2            & 58.7                            \\
\rowcolor[rgb]{0.831,0.831,0.831} \makecell[l]{MambaVLT*}   & 65.4         & 51.5          & 72.3            & 60.1                            \\ 
\hline
\multicolumn{5}{c}{\makecell{NL\&BBOX}}                                                                                                                                                                     \\ 
\hline
\makecell[l]{MambaVLT}                                      & 69.9         & 58.9          & 77.9            & 67.9                            \\
\rowcolor[rgb]{0.831,0.831,0.831} \makecell[l]{MambaVLT*}   & 70.2         & 59.1          & 79.0            & 68.6                             \\
\bottomrule
\end{tabular}
\end{table}
We introduce a new metric, success rate (SR), to align with the official experiments in the MGIT dataset. The prediction with the intersection over union $IoU$ that is higher than the threshold $\theta_{s}$ is regarded as a successful prediction. SR denotes the percentage of successfully tracked frames. According to Table \ref{sup:mgit}, in the NL and NL\&BBOX tasks, particularly the NL task, performance improves when natural language information is updated with the latest descriptions. This demonstrates that the state space memory is capable of modeling varying information.

\subsection{Efficiency Analysis}
% 分析模型的Flops PARAMS
% MambaAD 参考
\begin{table}[h]
\centering
\small
\setlength{\extrarowheight}{-5.5pt}
\caption{Efficiency comparison of state-of-the-art approaches. $x$ and $z$ denote the search region image and the template image.}
\label{sup:flop}
\begin{tabular}{l|ccc|cc} 
\toprule
Tracker  & $x$ & $z$ & \# $z$ & Params & FLOPs \\
\hline
\makecell[l]{JointNLT~\cite{zhou2023joint}} & 320    & 128      & 1           & 193 M  & \makecell{90 G}  \\
\makecell[l]{UVLTrack~\cite{ma2024unifying}} & 256    & 128      & 1           & 169 M  & 71 G  \\
\rowcolor[rgb]{0.831,0.831,0.831} MambaVLT & 256    & 128      & 3           & 149 M  & 70 G  \\
\bottomrule
\end{tabular}
\end{table}
\begin{figure}[H]
    \centering
    % \flushright
    \includegraphics[width=0.8\linewidth]{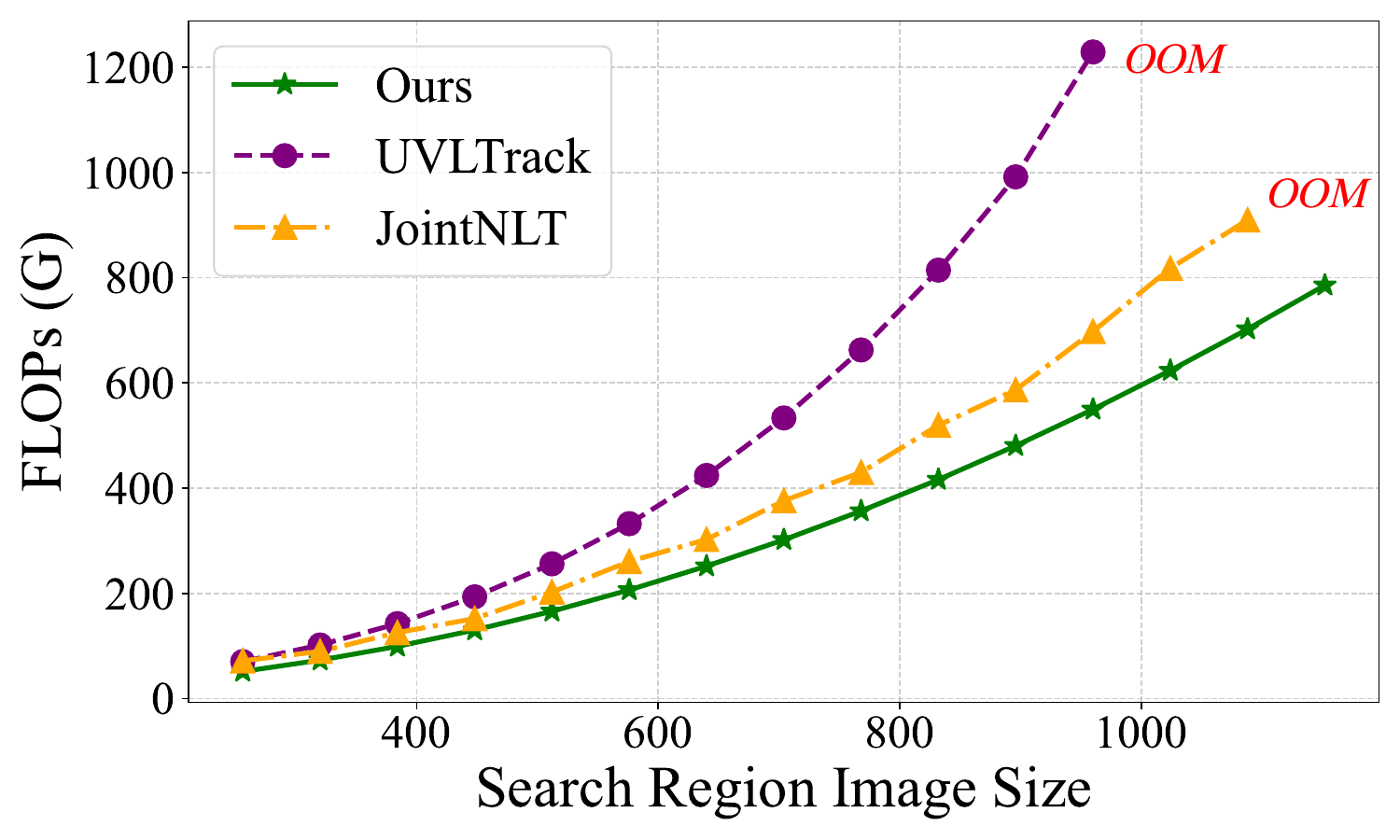}
    \caption{Computational complexity comparison with different search region image scales. OOM represents the computation cost is out of memory.
    }
    \label{sup:flopplot}
\end{figure}

\begin{figure*}
    \centering
    \includegraphics[width=0.7\linewidth]{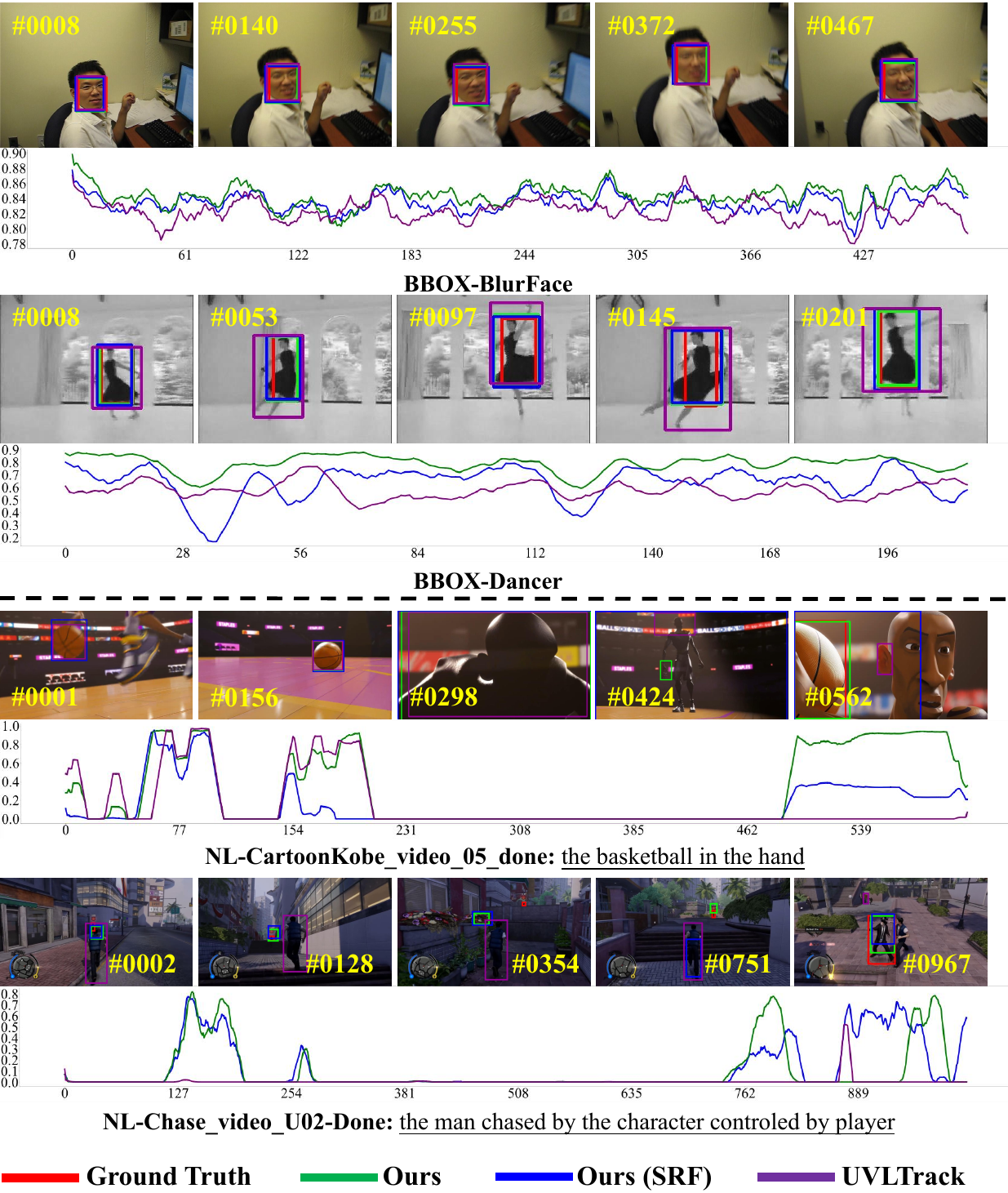}
    \caption{Effectiveness analysis of the time-evolving state space memory in BBOX and NL tasks. Under the \textbf{semi-reference-free} setting, the state space memory can still effectively extract and retain target features for accurate target localization compared to MambaVLT and UVLTrack using the standard tracking settings, validating the effectiveness of the state space memory.}
    \label{sup:vis_ss}
\end{figure*}
We employ the number of model parameters and floating-point operations (FLOPs) to evaluate the model size and computational complexity~\cite{liu2024vmamba,he2024mambaad}. As shown in Table \ref{sup:flop}, given the same search region size and language length, although we use three templates, the computational complexity of MambaVLT remains comparable to the models using a single template. Figure \ref{sup:flopplot} further investigates the trend of the computational complexity when scaling the search region size, while keeping other settings consistent with those in Table \ref{sup:flop}. As the search region size increases, the computational complexity of MambaVLT grows slowly, while that of UVLTrack shows a rapid quadratic growth trend. JointNLT reduces computational complexity by introducing the Swin Transformer~\cite{liu2021swin}.

\subsection{More Results of Semi-reference-free Tracking}
% 放三个任务下的srf结果
To analyze the effectiveness of state space memory, we design the semi-reference-free tracking paradigm, in which the reference data (language or initial bounding box) is used by tracker \textbf{only} in the first frame. From the second frame, the tracker needs to locate the target without explicitly using the reference data. The main challenge is extracting and memorizing target information based on the reference input in the first frame. We conduct the SRF-based experiments \textbf{without retraining}. As shown in Figure \ref{sup:vis_ss}, MambaVLT is able to track the target even without reference data after the first frame, demonstrating the effectiveness of the state space memory in target information retention.
\subsection{Qualitative Results}
% 放序列可视化结果
% 三个任务分别放可视化，这个放一些长序列可视化结果
% 仿照cite tracker分析鲁棒性 画表格显示AUC（这个可以放短序列的结果 选择template会在后面产生很大变化的？）

\begin{table*}[ht]
\centering
\setlength{\tabcolsep}{1pt}
\begin{tabular}{m{2cm}<{\centering} m{3.5cm}<{\centering} m{4cm}<{\centering} m{2cm}<{\centering} m{2cm}<{\centering} m{2cm}<{\centering} m{2cm}<{\centering}}
\hline
\makecell{\textbf{Task}} & \makecell{\textbf{Initial Frame}} & \makecell{\textbf{Language Description}} & \makecell{\textbf{Interference}} & \makecell{\textbf{UVLTrack}} & \makecell{\textbf{MambaVLT}} \\
\hline
BBOX & \includegraphics[width=0.15\textwidth]{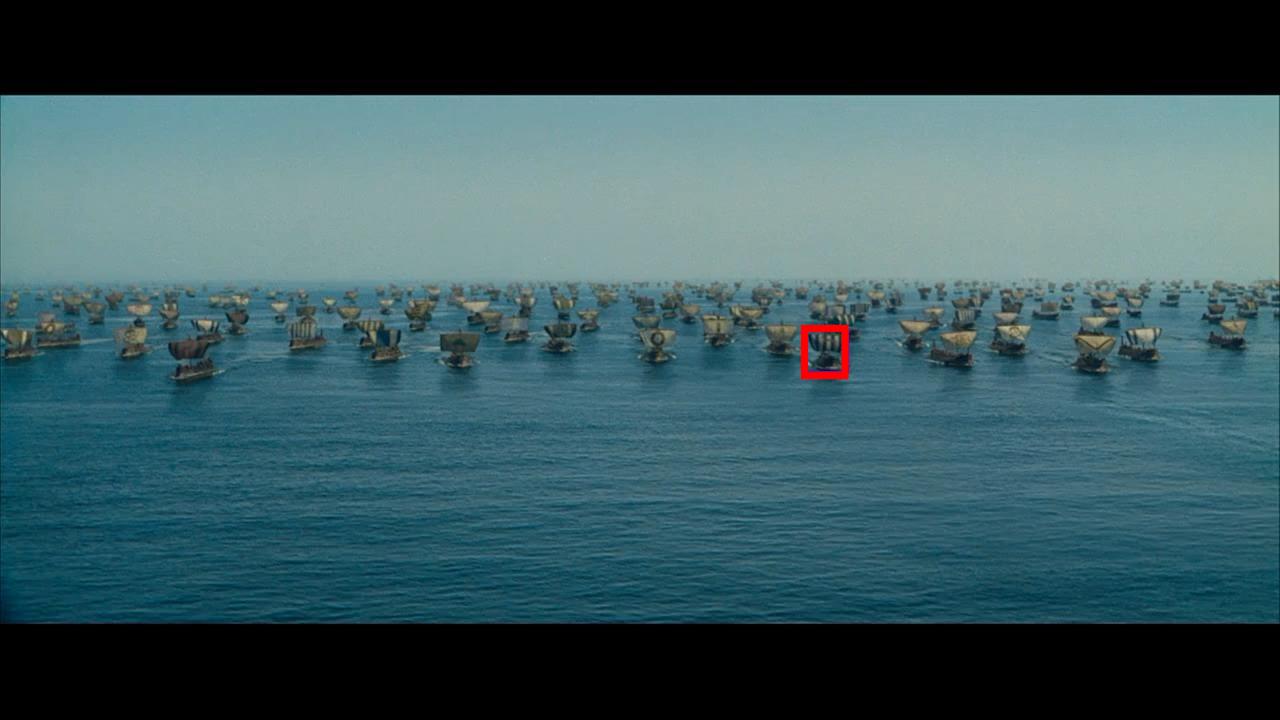} & - & Distractor & 56.1\%  & 70.2\%  \\
\hline
BBOX &\includegraphics[width=0.15\textwidth]{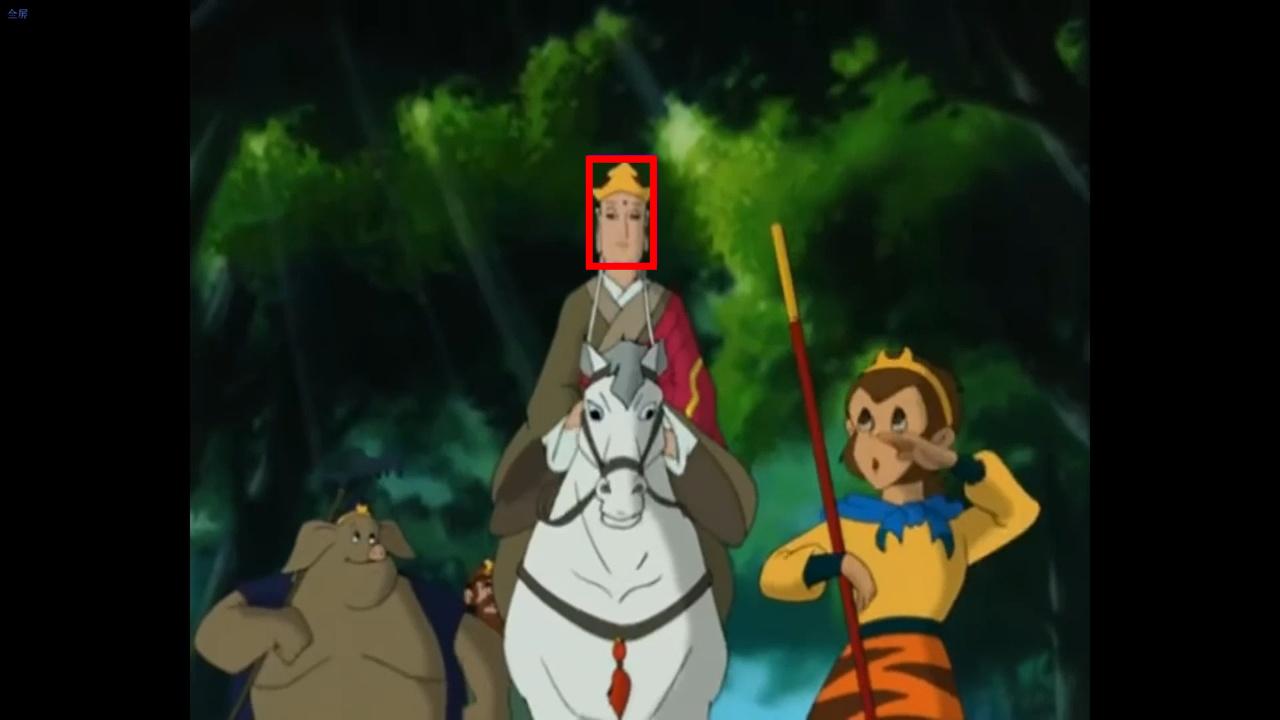} & - & Viewpoint Change & 64.2\% & 73.4\% \\
\hline
BBOX &\includegraphics[width=0.15\textwidth]{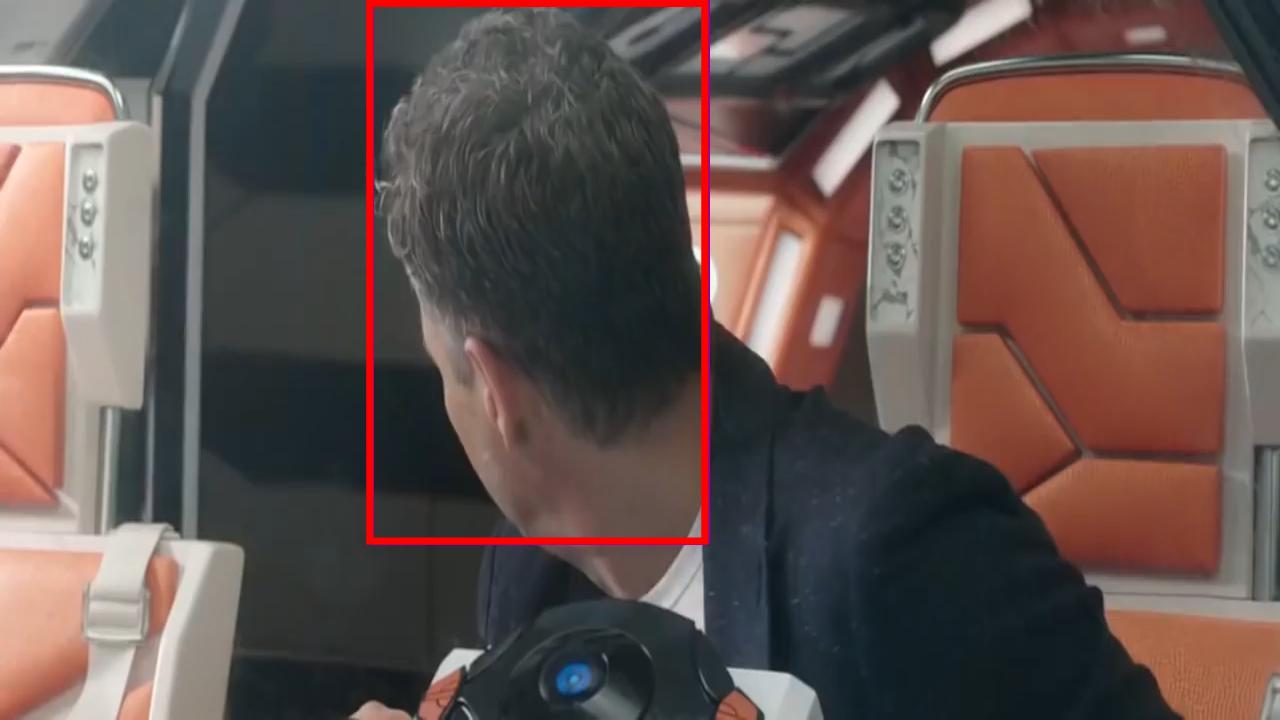} & - & Occlusion & 52.9\% & 66.1\% \\
\hline
NL & \includegraphics[width=0.15\textwidth]{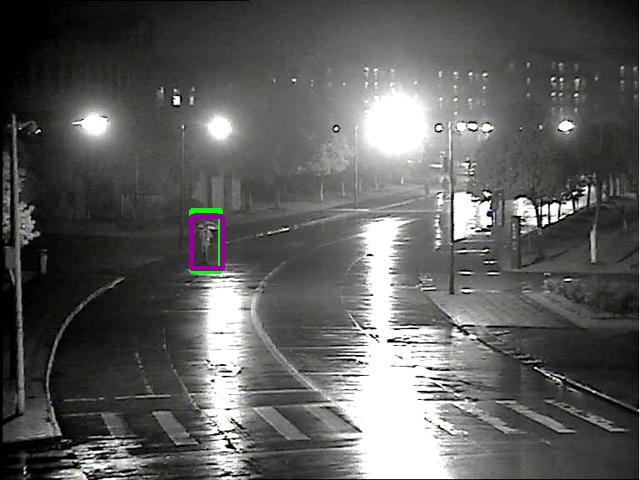} & we want to track a man holding an umbrella under street lamp & Low Light & 1.1\% & 66.3\% \\
\hline 
NL & \includegraphics[width=0.15\textwidth]{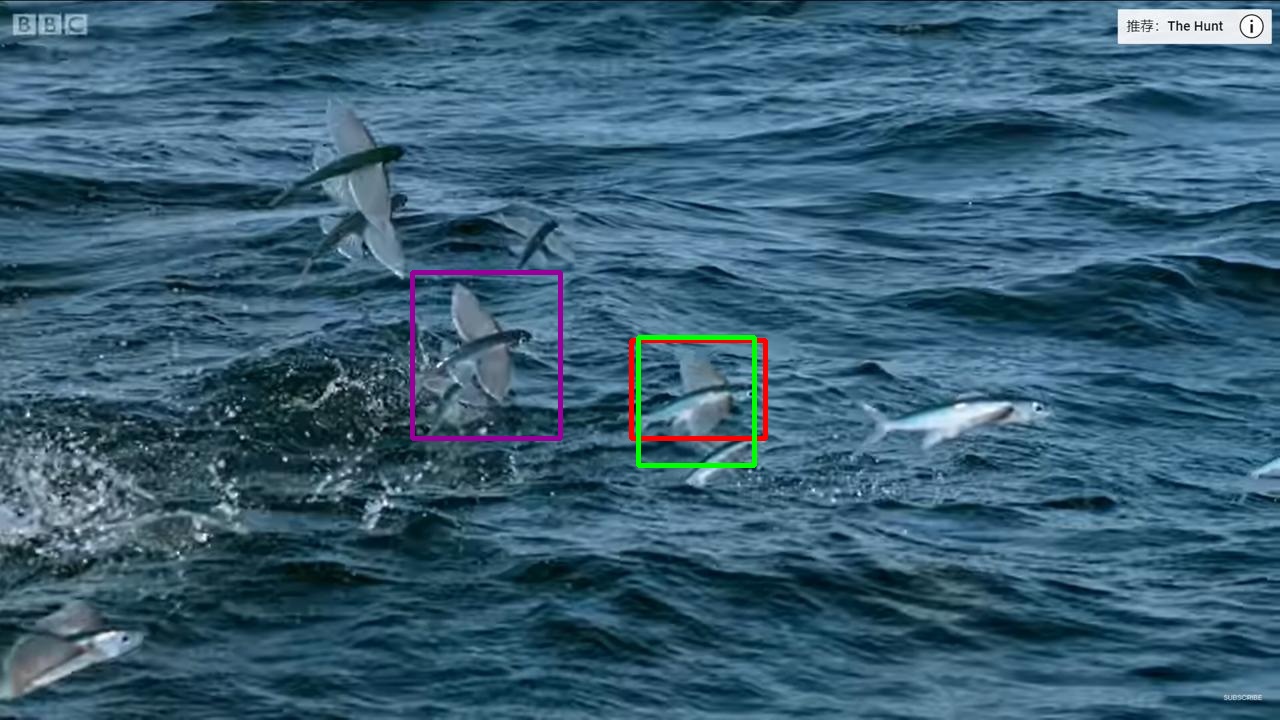} & the fourth fish from right to left & Distractor & 16.7\% & 42.5\% \\
\hline
NL & \includegraphics[width=0.15\textwidth]{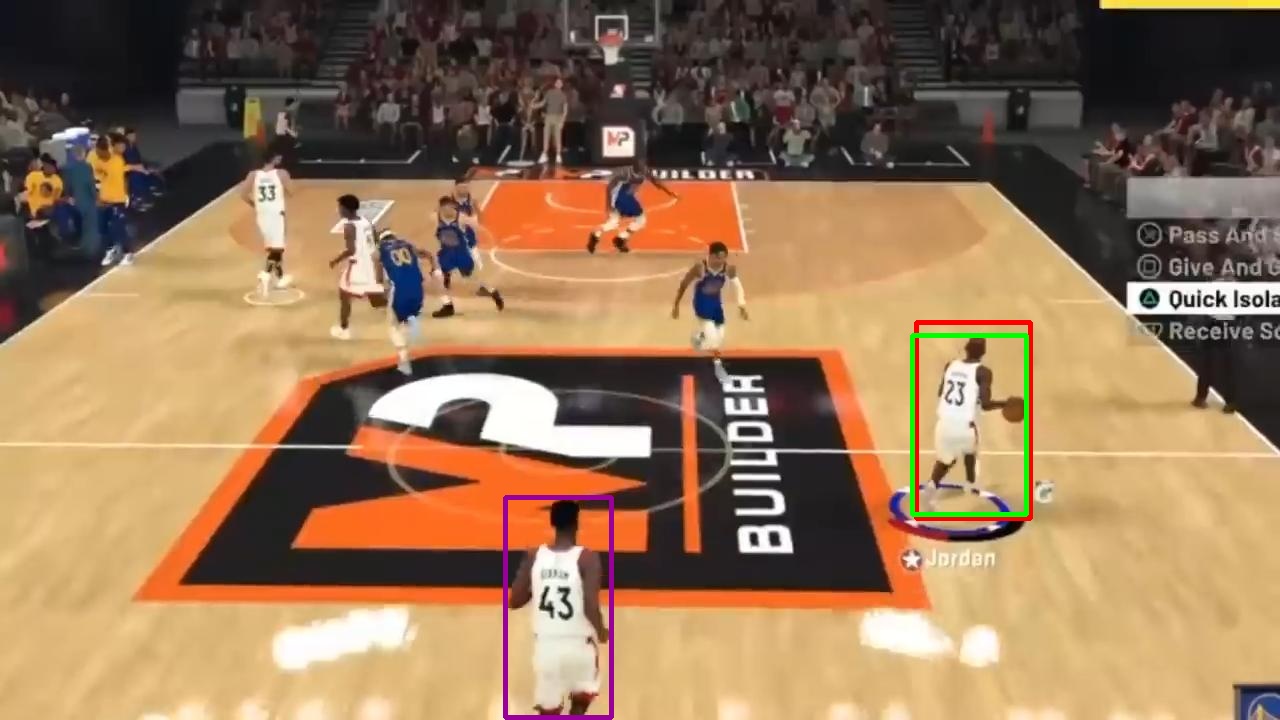} & the player wears white suit with twenty-three on this back  & Distractor & 4.0\% & 64.4\% \\
\hline
NL\&BBOX & \includegraphics[width=0.15\textwidth]{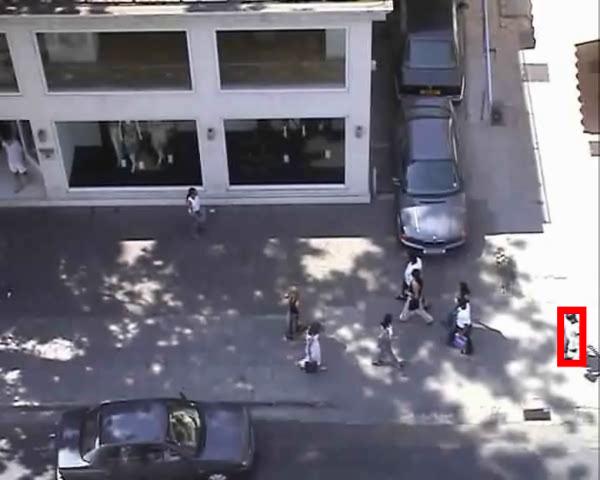} & the rightmost pedestrian in white & Distractor & 8.5\% & 75.4\% \\
\hline
NL\&BBOX & \includegraphics[width=0.15\textwidth]{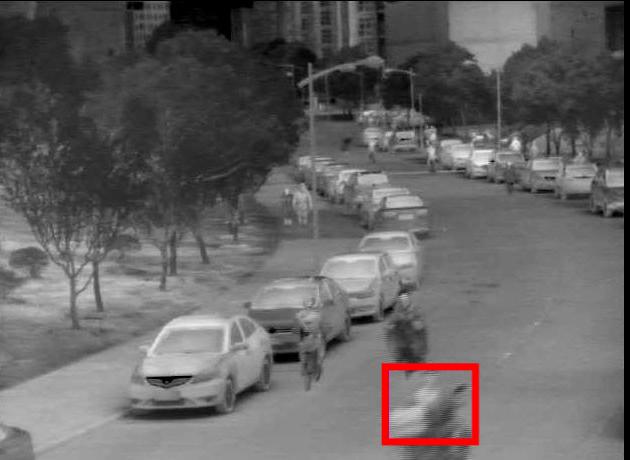} & the man on the bottom right corner  & Low Light & 16.3\% & 72.7\% \\
\hline
NL\&BBOX & \includegraphics[width=0.15\textwidth]{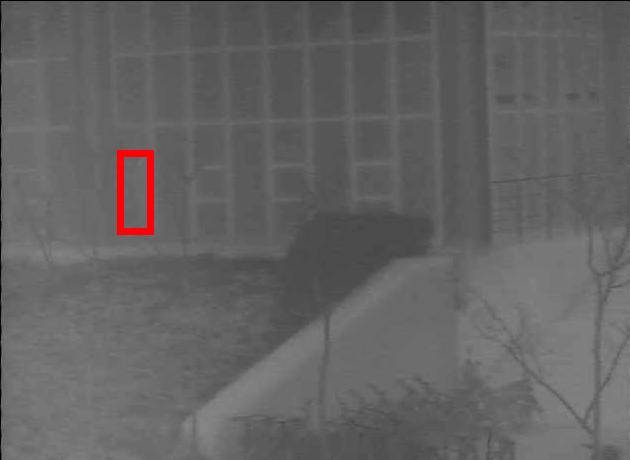} & the person on the corridor  & Occlusion & 13.4\% & 41.1\% \\
\hline

\end{tabular}
\caption{Robustness evaluation in terms of AUC score on several challenging sequences. It demonstrates that the MambaVLT significantly improves vision-language tracking performance by updating reference features adaptively in the cases where distractions exist between the target and reference information.}
\label{tab:tracker-performance}
\end{table*}

\begin{figure*}
    \centering
    \includegraphics[width=0.78\linewidth]{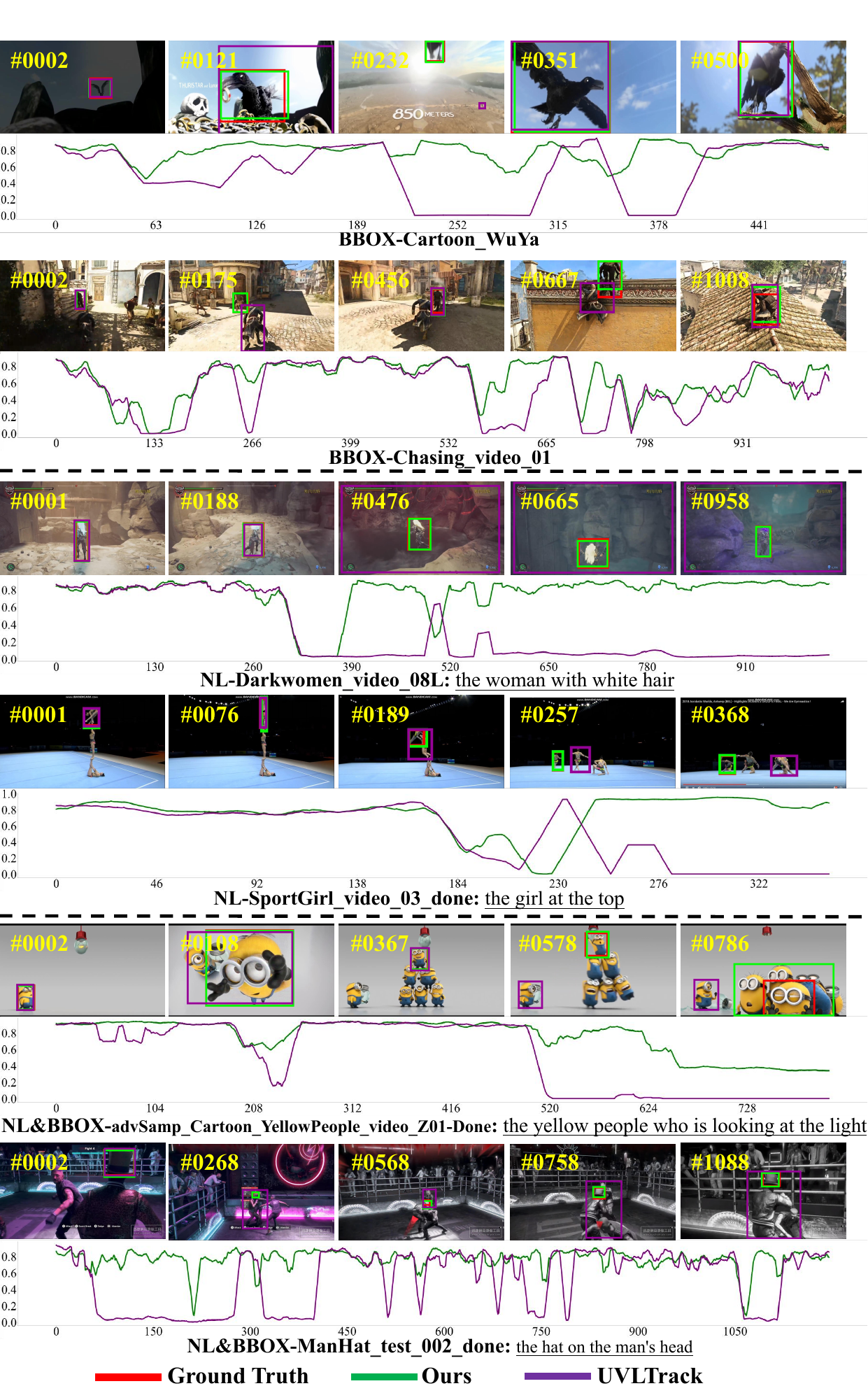}
    \caption{Visualized results of the MambaVLT and the UVLTrack method on six challenging sequences with \textbf{drastic changes}. Our MambaVLT performs well with the aid of the time-evolving state space memory for long-term target feature retention and adaptive reference feature update, while the UVLTrack with discrete context prompts struggles with these sequences.}
    \label{sup:vis_all}
\end{figure*}

Table \ref{tab:tracker-performance} presents the detailed results and the corresponding reference data of several sequences for robustness evaluation. The results indicate that MambaVLT has strong robustness against interference from the initial reference information, because our model can adaptively update the reference features and dynamically weigh multimodal information for modality selection. In Figure \ref{sup:vis_all}, we evaluate MambaVLT on six sequences characterized by drastic target variations. MambaVLT can still track the targets accurately, which demonstrates the introduction of time-evolving state space memory can help the model to retain long-term target features to update reference features for modeling long-term target variations adaptively. 

\end{document}